\documentclass[conference]{IEEEtran}
\usepackage{cite}
\usepackage{amsmath,amssymb,amsfonts}
\usepackage{algorithm}
\usepackage{algpseudocode}
\usepackage{graphicx}
\usepackage[normalem]{ulem}
\usepackage{textcomp}
\usepackage{enumitem}
\usepackage{arydshln} % Package for dashed and dotted lines
\usepackage[table]{xcolor} % Package for coloring table rows
\usepackage{subcaption}
\newcommand{\vect}[1]{\boldsymbol{#1}} % vector
\newcommand{\gpt}{\textsc{GPT3-1T}}
\newcommand{\vit}{\textsc{ViT}}
\newcommand{\nvs}{\textsc{NVS}}
\newcommand{\ib}{\textsc{IB}}
\newcommand{\hbm}{\textsc{HBM}}
\newcommand{\amp}{\textsc{A100}}
\newcommand{\hop}{\textsc{H200}}
\newcommand{\blackwell}{\textsc{B200}}
\newcommand{\SA}{\textsc{S/A}}
\newcommand{\LA}{\textsc{L/A}}
\newcommand{\LN}{\textsc{LN}}
\newcommand{\MLP}{\textsc{MLP}}
\newcommand{\SM}{\textsc{SM}}
\newcommand{\revision}[1]{{#1}}
\usepackage{hyperref}
\hypersetup{
    colorlinks=true,
    linkcolor=blue,
    filecolor=blue,
    urlcolor=blue,
    citecolor=blue
}
\begin{document}

% \title{Performance Insights and System Design Considerations for Large Transformer Models}
\title{Comprehensive Performance Modeling and System Design Insights for Foundation Models}

% \author{\IEEEauthorblockN{1\textsuperscript{st} Anonymous}}

\author{
\IEEEauthorblockN{1\textsuperscript{st} Shashank Subramanian}
\IEEEauthorblockA{
        \textit{Lawrence Berkeley National Laboratory} \\ 
        % \textit{Laboratory} \\
        Berkeley, CA, USA \\
        shashanksubramanian@lbl.gov
    }
\and
\IEEEauthorblockN{2\textsuperscript{nd} Ermal Rrapaj}
\IEEEauthorblockA{
        \textit{Lawrence Berkeley National Laboratory} \\ 
        % \textit{Laboratory} \\
        Berkeley, CA, USA \\
        ermalrrapaj@lbl.gov
    }
\and
\IEEEauthorblockN{3\textsuperscript{rd} Peter Harrington}
\IEEEauthorblockA{
        \textit{Lawrence Berkeley National Laboratory} \\ 
        % \textit{Laboratory} \\
        Berkeley, CA, USA \\
        pharrington@lbl.gov
    }
\and
\IEEEauthorblockN{4\textsuperscript{th} Smeet Chheda}
\IEEEauthorblockA{
        \textit{Stony Brook University, Dept. of CS} \\
        % \textit{University} \\
        Stony Brook, NY, USA \\
        schheda@cs.stonybrook.edu
    }
\and
\IEEEauthorblockN{5\textsuperscript{th} Steven Farrell}
\IEEEauthorblockA{
        \textit{Lawrence Berkeley National Laboratory} \\ 
        % \textit{Laboratory} \\
        Berkeley, CA, USA \\
        sfarrell@lbl.gov
    }
\and
\IEEEauthorblockN{6\textsuperscript{th} Brian Austin}
\IEEEauthorblockA{
        \textit{Lawrence Berkeley National Laboratory} \\ 
        % \textit{Laboratory} \\
        Berkeley, CA, USA \\
        baustin@lbl.gov
    }
\and
\IEEEauthorblockN{7\textsuperscript{th} Samuel Williams}
\IEEEauthorblockA{
        \textit{Lawrence Berkeley National Laboratory} \\ 
        % \textit{Laboratory} \\
        Berkeley, CA, USA \\
        swWilliams@lbl.gov
    }
\and
\IEEEauthorblockN{8\textsuperscript{th} Nicholas Wright}
\IEEEauthorblockA{
        \textit{Lawrence Berkeley National Laboratory} \\ 
        % \textit{Laboratory} \\
        Berkeley, CA, USA \\
         njwright@lbl.gov
    }
\and
\IEEEauthorblockN{9\textsuperscript{th} Wahid Bhimji}
\IEEEauthorblockA{
        \textit{Lawrence Berkeley National Laboratory} \\ 
        % \textit{Laboratory} \\
        Berkeley, CA, USA \\
        wbhimji@lbl.gov
    }
}

\maketitle

\begin{abstract}
Generative AI, in particular large transformer models, are increasingly driving HPC system design in science and industry. We analyze performance characteristics of such transformer models and discuss their sensitivity to the transformer type, parallelization strategy, and HPC system features (accelerators and interconnects). We utilize a performance model that allows us to explore this complex design space and highlight its key components. We find that different transformer types demand different parallelism and system characteristics at different training regimes. Large Language Models are performant with 3D parallelism and amplify network needs only at pre-training scales with reduced dependence on accelerator capacity and bandwidth. On the other hand, long-sequence transformers, representative of scientific foundation models, place a more uniform dependence on network and capacity with necessary 4D parallelism. Our analysis emphasizes the need for closer performance modeling of different transformer types keeping system features in mind and demonstrates a path towards this.  Our code is available as open-source at \cite{code}.

\end{abstract}


\section{Introduction}
Transformers \cite{Vaswani:2017}  have enabled state-of-the-art results across several disciplines including natural language processing (NLP)~\cite{Radford:2018,Devlin:2019,Colin:2020,Zhuang:2021,Du:2022}, 
computer vision (CV)~\cite{Kolesnikov:2021,Nouby:2021,Touvron:2021} and scientific machine learning (SciML)~\cite{chen2023fuxi,bodnar2024aurora,nguyen2023climax}. These models are expected to form the \emph{foundation} for machine learning models across various disciplines, owing to their impressive scaling properties with data and model sizes \cite{kaplan2020scaling}.
Large-language models (LLMs) are the most well-known foundation models such as GPT models~\cite{Brown:2020} with model sizes ranging to a trillion parameters~\cite{Narayanan:2021,Dash:2023} and are trained at large supercomputers at significant cost. 
For instance, the 1 trillion parameter Megatron GPT model was trained on 450 billion tokens, using 3072 Nvidia A100 GPUs, and requires 84 days~\cite{Narayanan:2021}.
%with a $50\%$ peak performance. 
Today, this cost is amplified by the need to train such models across multi-disciplinary domains in science---there is substantial effort to develop multiple foundation models for applications such as weather and climate modeling \cite{bodnar2024aurora,nguyen2023climax}, earthquake modeling \cite{sheng2023seismic}, fluid dynamics simulations \cite{mccabe2023multiple}, protein structure predictions \cite{jumper2021highly,lin2023evolutionary}, material sciences \cite{batatia2023foundation}, and more. Each domain introduces its own design considerations for transformer models, as well as unique input data scales (resolutions) and training methodologies.
This strongly motivates the need to theoretically analyze and understand the costs and bottlenecks associated with training different transformer types at scale and their relationship with the underlying system hardware. Our goal is to systematically present a framework for modeling the transformer performance and utilize this to explore optimal training strategies of the transformer at different system scales, transformer model regimes, and underlying system (hardware and network) characteristics. 

Building this {analytical performance model} is challenging because the design space is extremely large---one must model system configurations like device (GPUs) memory capacity and bandwidth, compute speeds, and (multi-bandwidth) network configurations, along with various parallelization strategies that grow exponentially with model and system size. The latter is especially complex because large transformers typically demand multiple forms of parallelism simultaneously, 
with each introducing its own set of trade-offs. Data parallelism is effective but insufficient for large models (and input resolutions) and not memory-efficient, tensor parallelism \cite{Shoeybi:2019,Narayanan:2021} reduces memory utilization at the cost of increased communication, and pipeline parallelism \cite{Narayanan:2021} can reduce communication costs at the expense of idle (bubble) times.
% In data parallelism \cite{what}, the batch of data is distributed across devices with the model \emph{shared} across all devices. For large models (or large input sequences/resolutions), due to the limited device memory capacity, model parallelism (tensor and pipeline parallelism) \cite{Shoeybi:2019,Narayanan:2021} is employed to partition the model (and inputs) along different dimensions. Each strategy comes with its own set of trade-offs. Increased tensor parallelism leads to lower memory utilization but increases communication time significantly and is strongly affected by the network configuration. Higher dimensional tensor parallelism \cite{Li:2023} can alleviate the communication volume to some degree. Pipeline parallelism may be used to reduce communication costs but it introduces idle time on the device (known as bubbles), and can drop performance efficiency significantly as well. Data parallelism may also introduce high communication times if it cannot be overlapped sufficiently behind compute. All these strategies closely depend on the transformer model hyperparameters as well. 
Overextending any one strategy can introduce substantial costs and they must be carefully balanced.
We present and utilize an analytic \emph{parameterized performance model} that searches the combinatorial design space 
% helps explore this design space analytically
and discovers this balance as a function of the deep learning model and system configuration.
% While heuristics and experimentation~\cite{Narayanan:2021,Li:2023,Dash:2023} help shed some light, it is impossible to search the design space through experimentation alone, and systematically discovering the right balance of parallelization strategies as a function of model hyperparameters and system configuration necessitates a \emph{parameterized performance model}. 
While our performance model is applicable to any accelerator system with tiers of network performance, we present detailed performance analysis for commonly used system architectures with current and near-future NVIDIA GPU hardware with high-bandwidth memory (\hbm) on device, a fast interconnect domain (compute node) with NVLink interconnect between GPUs (facilitated through NVSwitch \cite{nvswitch}), a slow interconnect domain (inter-node) InfiniBand (or SlingShot/ethernet) for GPUs, and communication collectives     through NCCL \cite{nccl}. 
%Our analysis is applicable to other architectures and networks as well with the assumption of equivalent components in those systems. 
Our contributions are as follows:
% \vspace{-0.5pt}
\begin{enumerate}[left=0pt, label=\arabic*.]
    \item 
    \textbf{Framework for building an AI performance model. }We systematically outline the strategy to model the various components of the transformer architecture. 
    We model the different operations (activation-weights, activation-activation matrix multiplies, vector operations) and highlight how parallelization strategies and other optimizations change the nature of these operations (arithmetic intensities, memory usage, communication times and other inefficiencies, see \S\ref{sec:methods}). 
    Defining a configuration as the parallelization strategy along with other possible optimizations, the framework identifies an optimal configuration through a brute-force search of all possible configurations and selects the one with minimum training time, making this exploration orders of magnitude faster than experimentation.
    \item
    \textbf{Assessment of varying needs of SciML and NLP models.}
    % We are interested in analyzing the different requirements of a SciML model and an NLP model at different training scales (large-scale pre-training and moderate-scale fine-tuning). 
     We assess two different transformer versions: \gpt ~(1 trillion parameter GPT3 model) representing an LLM foundation model and \vit, a long-sequence vision transformer that represents transformers in science, where long sequences are necessary to process high-resolution inputs, designed to capture crucial fine-scaled physical features \cite{willard2024analyzing,kurth2023fourcastnet,nguyen2023climax}. We then conduct the below analyses for both models independently, aiming to distinguish their training needs at different scales and across different systems.
    \item
    \textbf{Identifying optimal parallelism and training bottlenecks. }
    We include data parallelism, tensor parallelism (1D), and pipeline parallelism that are part of standard LLM training in our framework and demonstrate their individual trade-offs in training times (see Fig. \ref{fig:rationale}, \ref{fig:rationale-nvs}). We further include two different 2D versions of tensor parallelism (with additional sequence parallelism or SUMMA \cite{van1997summa} matrix multiplies) to demonstrate their impact on training the different model classes (see Fig. \ref{fig:rationale-nvs-2d}, \ref{fig:vit-par}). We show that while 1D tensor parallelism is performant for \gpt ~(see Fig. \ref{fig:gpt3-time}), 2D tensor parallelism is necessary for \vit ~and depicts different training bottlenecks than the 1D version (see Fig. \ref{fig:vit-par}). For both models, 2D tensor parallelism is optimal.
    \item
    \textbf{Performance impact of a fast network domain and GPU generations.}
    We analyze the impact of the size of NVSwitch (\nvs) that provides high bandwidth between all GPUs in that domain (node), with cross-domain (inter-node) slower InfiniBand (\ib). We show that placement of GPUs from different parallelization groups within \nvs ~domain introduces
    % subtle non-convexities in the landscape of the design-space with 
    different optimal configurations identified at different \nvs ~sizes (see discussions around Fig. \ref{fig:rationale-nvs}, \ref{fig:rationale-nvs-2d}). Further, we show that, depending on the type of model, \nvs ~size effects show up at different scales, lending to different requirements for pre-training and fine-tuning jobs. \gpt ~only requires large \nvs ~domain sizes at large training scale, needed for pre-training.
    %where realistic time-scales for training are in $\mathcal{O}$(months) and such performance benefits are crucial. 
    However, a moderate scale fine-tuning may not require this.
    We also notice that \hbm ~capacity is less important at scale for this model (see Fig. \ref{fig:overall-time}). The \vit ~however shows contrasting results with a more uniform dependence on \nvs ~domain sizes and \hbm ~capacity, owing to its massive sequence lengths (see Fig. \ref{fig:vit-time}). Both models show good performance with alternate low bandwidth/high capacity memory, which may help alleviate the dependence on \nvs ~through increased capacity. We also show the performance improvements with different GPU generations (\amp, \hop, \blackwell) for each model which can be attributed to increased tensor core and network bandwidth performance.
\end{enumerate}
% Our results emphasize the importance of the performance model and suggest a systematic way to explore optimizations at scale and understand the impact of systems on different model types. 

\section{Related Work}
\label{sec:related_work}
 We focus on analytical modeling and do not consider lower-level hardware simulation studies aimed at precise modeling of GPUs or the network since they are very costly for a full design space search.
 % Here we do not consider event-driven hardware modeling studies aimed at precise simulations of GPUs or network communications, but focus on recent work aimed at exploring all the dimensions of parallelism available for distributed training.
Our parallelization strategies are modeled based on tensor parallelism (1D \cite{Shoeybi:2019}, multi-dimensional~\cite{Xu:2023,Wang:2023,Bian:2021}), pipeline parallelism \cite{Narayanan:2021} and ZeRO \cite{Rajbhandari:2020}.
% 1D tensor parallelism was introduced in~\cite{Shoeybi:2019}. Subsequently, higher order tensor parallelisms were proposed to partition tensors in 2D~\cite{Xu:2023}, 2.5D~\cite{Wang:2023} and 3D~\cite{Bian:2021}. 
% Other optimized versions are present as well in Megatron and DeepSpeed code repositories \cite{codes} and we rely on these for our performance model. 
% Further strategies to reduce memory consumption include the ZeRO techniques~\cite{Rajbhandari:2020,Rajbhandari:2021} that use the data parallel devices to further partition gradients and optimizer states.
% which we include.
An early performance model exploring the design space is~\cite{Narayanan:2021} for training a version of \gpt. 
They do not conduct an automatic exploration of the design space but instead develop heuristic takeaways by focusing on a subset of the design space through analytical formulae. 
One of the first attempts at an automatic exploration and selection of parallelism strategies was by~\cite{Zheng:2022}, where given a cluster mesh, the package compiles the computation graph into a distributed sharded graph, but is not network topology aware. In~\cite{Li:2023}, they developed an open-source package that integrates an auto-parallelism feature that searches over the design space. Such auto-parallel approaches provide the user with a ready distributed training solution at the cost of hiding the rationale behind the specific strategy adopted. 
Calculon~\cite{Isaev:2023} was the first approach in building an in-depth, analytical performance model of both the LLM and the system for a high level and a \emph{unified} exploration of the design space of hardware and software.  Through the performance model, the authors identified new configurations that show better efficiencies for LLMs.  However, their model is restricted to LLMs and analysis fixed to a system type.
They only include 1D tensor parallelism (performant for LLMs), do not consider the effects of \nvs ~domains on \ib ~bandwidths and optimal strategies regarding placement of GPUs on the \nvs ~domain, and do not include details on the diverse components of the underlying performance model, making it challenging to expand upon.
% ---Calculon functions as a black-box.
% Further, they restrict their analysis to LLMs and a fixed system type. 
We build on the Calculon approach and expose the different performance modeling components along with several modifications that include \nvs ~domain size effects, GPU placement, 2D tensor parallelism variants (necessary for other model types) and conduct an extended analysis on different model types, GPU, and network characteristics.

\section{Methods}
\label{sec:methods}
Denoting $b$ as batch size, $l$ as sequence length, $e$ as embedding dimension, $f$ as the hidden dimension (typically $f = 4e$) and $h$ as the number of attention heads, the transformer \cite{Vaswani:2017} processes input tensor $X \in \mathbb{R}^{b \times l \times e}$ and predicts an output tensor of the same dimensions. We define $e_h = e / h$ as the head dimension. In NLP, the input represents a sequence of embedded tokens and in CV (and science) it represents a sequence of embedded image patches (or pixels). 
The transformer consists of repeated blocks that contain self-attention (\SA) and a multi-layer perceptron (\MLP) defined as:
\begin{align*}
    \tilde{\mathbf{X}} = \text{LN}(\mathbf{X}), \quad \mathbf{Y} &= \text{S/A}(\tilde{\mathbf{X}}) \\
    \tilde{\mathbf{Y}} = \text{LN}(\mathbf{Y}), \quad \mathbf{O} &= \text{MLP}(\tilde{\mathbf{Y}}),
\end{align*}
where \LN ~is LayerNorm and the above blocks repeated depth $d$ times. In \SA, the input tensor $\tilde{\mathbf{X}}$ is first projected to $\mathbf{Q},\mathbf{K},\mathbf{V}$ $\in \mathbb{R}^{b \times h \times l \times e_h}$ through learnable weights $\mathbf{W_Q}, \mathbf{W_K}, \mathbf{W_V}$ $\in \mathbb{R}^{e \times e}$, followed by the Logit-Attend (\LA) operation and a final projection with learnable weights $\mathbf{W_p}$ $\in \mathbb{R}^{e \times e}$:
\begin{align*}
    \mathbf{A}=\text{SM}(\mathbf{Q} \mathbf{K}^T) \in \mathbb{R}^{b \times h \times l \times l},  \quad  \mathbf{S} = &\mathbf{A}\mathbf{V} \in  \mathbb{R}^{b \times l \times e} \\
    \mathbf{Y} =  \mathbf{S}\mathbf{W_p} \in \mathbb{R}^{b \times l \times e}.
\end{align*} 
\SM ~denotes Softmax. We note that the first two operations (\LA) involve only activation maps, and do not introduce weights. After a subsequent \LN, $\tilde{\mathbf{Y}}$ is passed to the \MLP ~defined as:
\begin{align*}
    \mathbf{O} = \text{GeLU}(\mathbf{Y} \mathbf{W_1} + \vect{b_1}) \mathbf{W_2} + \vect{b_2} \in \mathbb{R}^{b \times l \times e},
\end{align*}
with learnable weights $\mathbf{W_1} \in \mathbb{R}^{e \times f}$, $\mathbf{W_2} \in \mathbb{R}^{f \times e}$, learnable bias $\vect{b_1} \in \mathbb{R}^f$, $\vect{b_2} \in \mathbb{R}^e$, and output tensor $\mathbf{O}$. 
%
% The transformer blocks are repeated for depth $d$ times. 
Dropout layers are additionally present but we omit them here for brevity. 
For large data ($b$) and/or model dimensions ($l, e, h$), parallelizing the transformer is necessary. Assuming a grid of $n = n_d \times n_t \times n_p$ GPUs, this amounts to finding an optimal allocation of these GPUs to partition the above dimensions. At the highest level, $b$ is partitioned across $n_d$ GPUs ({data} parallelism, but may be subsumed into tensor parallelism),
$(l, e, h)$ are partitioned across $n_t$ GPUs ({tensor} parallelism) and $d$ (depth) is partitioned across $n_p$ GPUs ({pipeline} parallelism). The tensor parallel GPU group $n_t$ can be further split through a multi-dimensional array of GPUs ($n_t = n_1 \times n_2 \times \cdots$) for better performance. Each of these parallel strategies come with their own set of performance benefits and bottlenecks with several non-trivial factors at play that depend closely on the underlying system. In order to expose these trade-offs, identify an optimal parallelization strategy constrained by the system, and understand the effects of different system and model types on training times, we start with an analytical and parameterized performance model. 

\subsection{Parameterized Performance Model}
\label{sec:perf_model}
Given $n$ GPUs, the transformer model architecture, a global batch size $b$, and system characteristics (hardware, network), we compute the minimum theoretical time it takes to complete a forward and backward pass of the model to get an optimal training time estimate. We do this in three stages:

\begin{enumerate}[label=(\textbf{S\arabic*})] 
    \item We systematically count the total FLOPs (floating point operations), amount of memory accessed from \hbm ~and communication volume for {every} major operation in the transformer, as well as the amount of memory consumed in \hbm ~to hold intermediate activation maps and weights---these depend on the {parallelization strategy}.
    \item Given the above counts, we then compute the theoretical time it takes to complete a forward and backward pass for each layer---this depends on the underlying {system characteristics}. The total iteration time for the transformer is the sum of all individual layer times.
    \item Finally, we search over all possible parallelization configurations, given $n$ and $b$, to identify the optimal one with minimum training time with the only constraint that the model fits on the \hbm ~capacity. 
\end{enumerate}
\revision{Below, we describe the key components of these stages.
We start by counting the FLOPs and memory accesses for the matrix multiply primitive, given the shapes of the input and output tensors. Next, we outline how these computations change across different parallelization strategies, which generally amount to distributed matrix multiplies. In particular, we describe every operation of the transformer along with its input and output tensor shapes, showing how they are partitioned under various parallelization strategies. We also count the communication volume in bytes, highlighting how it depends on the partitioning of tensors and the communication collectives used, which we explain in depth.
We then briefly discuss the role of additional components, such as pipelining---primarily a scheduling challenge rather than distributed matrix multiplication---and optimizer partitioning, as these have a particular impact on memory consumption. Once we have determined the counts for FLOPs, memory accesses, and communication volume, we demonstrate how to convert these into compute and communication times using an analytical runtime model. Finally, we describe our solver, which explores all possible configurations (i.e., all possible ways to use $n$ GPUs for partitioning the tensors in the model) to minimize the overall iteration time.
}

\medskip \noindent \textbf{(S1) Counting FLOPs and bytes. }
\revision{
Most transformer operations heavily rely on matrix multiplication $\mathbf{C} = \mathbf{A}\mathbf{B}$ where $\mathbf{C} \in \mathbb{R}^{m \times n}, \mathbf{A} \in \mathbb{R}^{m \times k}, \mathbf{B} \in \mathbb{R}^{k \times n}$. The total FLOPs in this operation is $\lambda_f = (2k - 1)mn$. The memory accessed from \hbm ~is $\lambda_m = 2(mk + kn + mn)$ bytes (corresponding to each tensor), assuming FP16 precision (we consider mixed precision training in our model). 
Similar expressions can be derived for \LN, \SM, GELU, and Dropout, which are simpler than matrix multiplication.
We also count communication volume in bytes based on the underlying collective operation.}
% Assuming peak hardware FLOPs $\lambda_{fh}$ and peak hardware memory bandwidth from \hbm ~as $\lambda_{mh}$, roofline performance dictates peak performance time for the operation as $\max (\lambda_{f}/\lambda_{fh}, \lambda_{m}/\lambda_{mh})$. 
% We can add communication time based on $V$ bytes of communication volume $t_\text{comm}$ 
% (see \S\ref{sec:methods}) 
% to this time to get the final estimate of the operation time. Depending on the operation, $t_\text{comm}$ may be overlapped with compute or exposed. 
% When operations are fused, as in \LA ~where the attention matrix ($\mathbf{A}$), \SM ~and attend operation to compute $\mathbf{S}$ are fused to a single operation, the amount of bytes counted only depend on the inputs to the fused operation and no intermediates (which increases the arithmetic intensity).
% }

\medskip \noindent \textbf{(S1) Fused Operations. }
The \SA ~layer is unique to the transformer because it involves large activation-activation computations (i.e., without learnable weights) that are also batched (many-to-many operations) \cite{kao2023flat}---this makes the operation fundamentally \emph{memory-bound}. Further, large intermediate activation maps (typically stored for backward pass) in $\mathbb{R}^{b \times h \times l \times l}$ puts significant memory pressure on the \hbm. We assume the \textsc{FlashAttention} \cite{dao2022flashattention} formulation that fuses \LA ~and re-computes intermediate activation maps in the backward pass (expending more FLOPs). This can bring \SA ~into a \emph{compute-bound} regime. For fused \LA, where the attention matrix ($\mathbf{A}$), \SM ~and attend operation for $\mathbf{S}$ are fused, the amount of bytes counted only depend on the inputs to the fused operation and no intermediates (increasing the arithmetic intensity).

\medskip \noindent \textbf{(S1) Tensor Parallel. }The standard form of tensor parallelism (TP) is \emph{1D} TP \cite{Shoeybi:2019}.
\begin{table}[htbp]
  \centering
  \rowcolors{2}{gray!25}{white} % Alternating row colors
  \begin{tabular}{c|c|c|c}
    % \hline
    Operation & Partitioned Tensor Shapes & Type &  Vol \\
    \hline
    \multicolumn{4}{|c|}{\textbf{1D TP over $n_t$ GPUs}} \\
    \hline
    \multicolumn{4}{c}{\textit{SA}} \\
    \hline
    $\tilde{\mathbf{X}} = \text{LN}(\mathbf{X})$   & $\tilde{\mathbf{X}}: (b,l,e)$, $\mathbf{X}: (b,\frac{l}{n_t},e)$,  & $\mathcal{AG}$ & $ble$ \\
    $\mathbf{Q} = \tilde{\mathbf{X}} \mathbf{W_Q}$   & $\mathbf{Q}: (b,\frac{h}{n_t},l,e_h)$, $\mathbf{W_Q}: (e,\frac{e}{n_t})$,  & - & 0 \\
    $\mathbf{A}=\mathbf{Q} \mathbf{K}^T$   & $\mathbf{A}: (b,\frac{h}{n_t},l,l)$, $\mathbf{K}: (b,\frac{h}{n_t},l,e_h)$  & - & 0 \\
    $\mathbf{S} = \mathbf{A}\mathbf{V}$   & $\mathbf{S}: (b,\frac{h}{n_t},l,e_h)$, $\mathbf{V}: (b,\frac{h}{n_t},l,e_h)$  & - & 0 \\
    $\mathbf{Y} =  \mathbf{S}\mathbf{W_p}$   & $\mathbf{Y}: (b,\frac{l}{n_t},e)$, $\mathbf{W_p}: (\frac{e}{n_t},e)$  &  $\mathcal{RS}$ & $ble$ \\
    \hline
    \multicolumn{4}{c}{\textit{MLP}} \\
    \hline
    $\tilde{\mathbf{Y}} = \text{LN}(\mathbf{Y})$   & $\tilde{\mathbf{Y}}: (b,l,e)$, $\mathbf{Y}: (b,\frac{l}{n_t},e)$,  &  $\mathcal{AG}$ & $ble$ \\
    $\mathbf{Z} = \tilde{\mathbf{Y}}\mathbf{W_1}$   & $\mathbf{Z}: (b,l,f/n_t)$, $\mathbf{W_1}: (e,\frac{f}{n_t})$  & - & 0 \\
    $\mathbf{X} = \mathbf{Z} \mathbf{W_2}$   & $\mathbf{X}: (b,\frac{l}{n_t},e)$, $\mathbf{W_2}: (\frac{f}{n_t},e)$  &  $\mathcal{RS}$ & $ble$ \\
  \end{tabular}
  \caption{\emph{1D TP: Tensor shapes, communication collective and volume (\emph{Vol}: total bytes transferred per GPU) for different operations. $\mathcal{AG}$ is AllGather and $\mathcal{RS}$ is ReduceScatter. $\mathbf{K}, \mathbf{V}$ follow $\mathbf{Q}$. Communication volume does not scale with $n_t$.}}
  \label{tab:1D}
\end{table}
Here, a 1D array of $n_t$ GPUs are used to partition the weight matrices (in row-parallel or column-parallel fashion) as well as the sequence length $l$. We show the different transformer operations with partitioned tensors using 1D TP and the associated communication collectives and volume in Tab. \ref{tab:1D}. The only communication collectives are AllGather $\mathcal{AG}$ and ReduceScatter $\mathcal{RS}$ with communication volume $ble$. Note that while the volume is fixed w.r.t $n_t$, it depends on $b$ which is closely connected to data (and pipeline) parallelism. In Tab. \ref{tab:1D}, we see that some tensors ($\tilde{\mathbf{X}}$, $\tilde{\mathbf{Y}}$) are replicated across the $n_t$ GPUs in 1D TP. For large $l$ and $e$, this might place extensive memory pressure and \SA ~still retains $\mathcal{O}(l^2)$ time complexity.
%since the $n_t$ GPUs are only used to partition the heads (making \LA ~embarrassingly parallel). 

We could instead also use a 2D grid of GPUs $n_t = n_1 \times n_2$ to partition each tensor, also known as context parallelism \cite{megatroncolde} or \emph{2D} tensor parallelism. We show the operations with 2D TP in Tab. \ref{tab:2D}. Here, $l$ is further partitioned in the orthogonal $n_2$ GPU group, reducing memory pressure, and incurs two additional $\mathcal{AG}$ for the tensors $\mathbf{K}, \mathbf{V}$. Further, all communication volumes scale by number of GPUs in their group, lending better scalability. We note that all weight tensors are \emph{shared} (redundant memory) in the $n_2$ group.
\begin{table}[htbp]
  \centering
  \rowcolors{2}{gray!25}{white} % Alternating row colors
  \begin{tabular}{c|c|c|c}
    % \hline
    Operation & Partitioned Tensor Shapes & Type &  Vol \\
    \hline
    \multicolumn{4}{|c|}{\textbf{2D TP over $n_1 \times n_2$ grid of GPUs}} \\
    \hline
    \multicolumn{4}{c}{\textit{SA}} \\
    \hline
    $\tilde{\mathbf{X}} = \text{LN}(\mathbf{X})$   & $\tilde{\mathbf{X}}: (b,\frac{l}{n_2},e)$, $\mathbf{X}: (b,\frac{l}{n_1n_2},e)$,  & $\mathcal{AG}$ & $b\frac{l}{n_2}e$ \\
    $\mathbf{Q} = \tilde{\mathbf{X}} \mathbf{W_Q}$   & $\mathbf{Q}: (b,\frac{h}{n_1},\frac{l}{n_2},e_h)$, $\mathbf{W_Q}: (e,\frac{e}{n_1})$,  & - & 0 \\
    $\mathbf{A}=\mathbf{Q} \mathbf{K}^T$   & $\mathbf{A}: (b,\frac{h}{n_1},\frac{l}{n_2},l)$, $\mathbf{K}: (b,\frac{h}{n_1},l,e_h)$  & $\mathcal{AG}$ & $bl\frac{e}{n_1}$ \\
    $\mathbf{S} = \mathbf{A}\mathbf{V}$   & $\mathbf{S}: (b,\frac{h}{n_1},\frac{l}{n_2},e_h)$, $\mathbf{V}: (b,\frac{h}{n_1},l,e_h)$  & $\mathcal{AG}$ & $bl\frac{e}{n_1}$ \\
    $\mathbf{Y} =  \mathbf{S}\mathbf{W_p}$   & $\mathbf{Y}: (b,\frac{l}{n_1n_2},e)$, $\mathbf{W_p}: (\frac{e}{n_1},e)$  &  $\mathcal{RS}$ & $b\frac{l}{n_2}e$ \\
    \hline
    \multicolumn{4}{c}{\textit{MLP}} \\
    \hline
    $\tilde{\mathbf{Y}} = \text{LN}(\mathbf{Y})$   & $\tilde{\mathbf{Y}}: (b,\frac{l}{n_2},e)$, $\mathbf{Y}: (b,\frac{l}{n_1n_2},e)$,  &  $\mathcal{AG}$ & $b\frac{l}{n_2}e$ \\
    $\mathbf{Z} = \tilde{\mathbf{Y}}\mathbf{W_1}$   & $\mathbf{Z}: (b,\frac{l}{n_2},\frac{f}{n_1})$, $\mathbf{W_1}: (e,\frac{f}{n_1})$  & - & 0 \\
    $\mathbf{X} = \mathbf{Z} \mathbf{W_2}$   & $\mathbf{X}: (b,\frac{l}{n_1n_2},e)$, $\mathbf{W_2}: (\frac{f}{n_1},e)$  &  $\mathcal{RS}$ & $b\frac{l}{n_2}e$ \\
    % \hline
    % \multicolumn{4}{|c|}{\textbf{2D: TP}} \\
    % \hline
    % % \hline
  \end{tabular}
  \caption{\emph{2D TP: Tensor shapes, communication collective and volume (\emph{vol}: total bytes transferred per GPU) for different operations. $\mathbf{K}, \mathbf{V}$ follow $\mathbf{Q}$. Communication volume scales with one GPU dimension. }}
  \label{tab:2D}
\end{table}
We can further improve memory pressure by using the {SUMMA}~\cite{van1997summa} algorithm for all activation-weight operations (no shared weights). For details on SUMMA, see Appendix \S\ref{sec:appendix_additional_methods}. The communication collectives change to two Broadcasts $\mathcal{B}$ per matrix multiply with communication volume scaling with both $n_1$ and $n_2$ (see Tab. \ref{tab:2D-summa}). Though the scaling of communication volume and memory is better than 2D TP, the actual volume is higher due to both activation maps and weights being transferred (see $V_1,V_2,V_3$ in Tab. \ref{tab:2D-summa}). Hence, depending on the relative sizes of $l,e,f$, a large amount of GPUs may be needed for reasonable communication volumes.
However, SUMMA also introduces more communication overlaps that may reduce with more partitioning (see Appendix \S\ref{sec:appendix_additional_methods} discussion on overlaps).

\medskip \noindent \textbf{(S1) Pipeline Parallel. }
The model can be partitioned in the depth $d$ dimension using $n_p$ pipeline parallel (PP) orthogonal GPUs. We assume the \emph{1F1B} non-interleaved pipeline schedule here. Here, a batch $b$ is split into $m$ microbatches of microbatch size $b_m = b/m$ to reduce idle time in GPUs \cite{Narayanan:2021}. Communication is point-to-point $\mathcal{P}2\mathcal{P}$ of volume $mb_mle/n_t$ (activation maps for $m$ microbatches). We do not assume overlapping this communication with compute (in \S\ref{sec:results}, we show this communication time is small and, hence,  a reasonable assumption)

\medskip \noindent \textbf{(S1) Data Parallel and Optimizer. }
We can use $n_d$ orthogonal data parallel (DP) GPUs to partition $b_m$ to $b_m/n_d$ for all devices. For large weight matrices, it is also common to distribute the optimizer states amongst $n_d$ GPUs. \revision{Assuming FP16 computations, 
this amounts to $12/n_d$ bytes of memory per parameter, assuming Adam optimizer}
% $2w$ bytes of total weight memory, and Adam optimizer ($12w$ bytes of optimizer states), this reduces the gradient and optimizer footprint to $14w/n_d$ 
\cite{Rajbhandari:2020}.
The forward pass is embarrassingly parallel and the backward pass incurs a $\mathcal{RS}$ and $\mathcal{AG}$ of the weight gradients and weights, respectively. We assume that gradient accumulation occurs over $m$ microbatches (no communication) and the $\mathcal{RS}$ is overlapped with the backward pass of the last microbatch and $\mathcal{AG}$ is overlapped with the forward pass of the first microbatch after the pipeline flush. Note that, in $2D$ TP, the weight gradients need an additional reduction across $n_2$ GPUs and we assume it to be scheduled simultaneously with the DP $\mathcal{RS}$ and $\mathcal{AG}$ (see Appendix \S\ref{sec:appendix_additional_methods}).
% $\mathcal{RS}$ amongst $n_2$ GPUs that can be implemented simultaneously with the DP $\mathcal{RS}$ (see Appendix \S\ref{sec:appendix_additional_methods}). 
At the end, we have $n = n_1 \times n_2 \times n_p \times n_d$ grid of GPUs for full 4D parallelism.

\medskip \noindent \textbf{(S2) Memory Used on HBM. }Assuming mixed precision training, in addition to the weights and gradients ($2$ bytes per parameter for each) and optimizer states ($12/n_d$ bytes per parameter), every operation stores intermediate activation maps (needed for backward pass, see Appendix \S\ref{sec:appendix_additional_methods}) for $m$ microbatches. With \textsc{FlashAttention}, the intermediate $\mathbf{A}$ are not stored but recomputed. The \emph{1F1B} pipeline schedule further reduces memory by storing $n_p$ microbatches (rather than $m$) since the schedule decreases the total number of in-flight microbatches for backward pass \cite{Narayanan:2021}.

\medskip \noindent \textbf{(S2) Computation Time. }
\revision{
We use the simple roofline model \cite{williams2009roofline} to convert FLOPs and memory accesses into computation time. 
Assuming the matrix multiply primitive (with similar expressions for non-matrix multiply operations) with $\lambda_f$ FLOPs and $\lambda_m$ memory accesses as well as hardware peak  FLOP rate $\lambda_{fh}$ and peak hardware memory bandwidth from \hbm ~as $\lambda_{mh}$, roofline performance dictates peak performance time for the operation as $\max (\lambda_{f}/\lambda_{fh}, \lambda_{m}/\lambda_{mh})$. We assume tensor core hardware FLOPs (in GPUs) for the matrix operations and vector hardware FLOPs for the others.
We can add communication time based on $V$ bytes of communication volume $t_\text{comm}$ 
% (see \S\ref{sec:methods}) 
to this time to get the final estimate of the operation time. Depending on the operation, $t_\text{comm}$ may be overlapped with compute or exposed. 
}

\medskip \noindent \textbf{(S2) Communication Time. }
We assume two networks--one fast network through NVSwitch (\nvs) for GPUs within a node with $(\alpha_f, \beta_f)$ as the network latency and bandwidth and one slow network through InfiniBand (\ib) across nodes with latency and bandwidth of $(\alpha_s, \beta_s)$. We assume $\mathcal{RS},\mathcal{AG}$ happen using the ring algorithm. While the \ib ~bandwidth $\beta_s$ is typically much smaller, NCCL can employ multiple rings, proportional to number of NICs (network-interface cards) per node $n_\text{NIC}$, to simultaneously complete the collectives\cite{ncclgtc}---this effectively increases the IB bandwidth to $n_\text{NIC}\beta_s$. Assuming $n_\text{NVS}$ GPUs per node (or per \nvs ~domain) and a total of $n$ GPUs for the communication collective, we model the time $t_{\text{comm}}$ for an $\mathcal{AG}$ of a total of $V$ bytes of volume per GPU as:
\begin{align*}
    t_{\text{latency}} &= \alpha_s\left(\frac{n}{n_\text{NVS}} - 1\right) + \alpha_f \left(n - \frac{n}{n_\text{NVS}} \right),\\
    t_{\text{comm}} &= t_{\text{latency}} + \frac{(n-1)}{n}\max{(\frac{V}{n_\text{NIC}\beta_s}, \frac{V}{\beta_f})}.
\end{align*}
Here, we have followed the theoretical time expressions for a single network given by NCCL performance models \cite{nccltests}. Typically, $n_\text{NIC}$ is equal (or proportional) to the \nvs ~domain size $n_\text{NVS}$. Hence, the effective bandwidth of communication $\max{(n_\text{NIC}\beta_s,\beta_f)}$ is eventually constrained by $\beta_f$ for large \nvs ~domains. Similar expressions can be derived for the other collectives. We empirically verify that these expressions model the communication time reasonably in a dual network system through NCCL tests (see Appendix \S\ref{sec:appendix_additional_results}).

\medskip \noindent \textbf{(S2) Pipeline Bubble Time. } Additional idle time is incurred in pipeline bubbles and is modeled as $t_{\text{bubble}} = (n_p - 1) (t_f + t_b)$ with $t_f$ and $t_b$ as the time it takes to complete forward and backward pass of one microbatch, respectively \cite{Narayanan:2021}.

\medskip \noindent \textbf{(S3) Optimal Configuration. }Given $n$ GPUs and a global batch size $b$, we identify an optimal configuration through combinatorial optimization---searching through all possible configurations and choosing a feasible configuration with minimum time, where feasibility is defined as having the ability to fit in \hbm. 
The configurations include the following:
\begin{enumerate}
    \item \textbf{Parallelization and microbatch size configurations}: \revision{These are contained in $(b_m,n_1,n_2,n_p,n_d)$. To get all possible configuration, we simply decompose $n = n_1 \times n_2 \times n_p \times n_d$ by sweeping through all possible factors. We discard factors that do not divide the tensor they are partitioning evenly. For example, if $n_2$ GPUs are used to partition the sequence length $l$ as $l/n_2$, then $n_2$ must divide $l$. Similarly, $n_d$ (data parallelism) must divide the global batch size, the microbatch size $b_m$ must divide the local batch size, and $n_p$ (pipeline parallelism) must divide the model depth. 
    }
    \item \textbf{GPU assignment configurations}: 
    \revision{
    These are specified by $(n_{\text{\nvs1}},n_{\text{\nvs2}},n_{\text{\nvs p}},n_{\text{\nvs d}})$,
    where $n_{\text{NVSi}}$ is the number of GPUs in the \nvs ~domain for the $i$-th GPU group. For example, $n_1 = 32$ with $n_{\text{\nvs1}} = 4$ indicates that groups of $4$ GPUs are on the \nvs ~domain (or equivalently $4$ GPUs per node are used across $8$ nodes). 
    Searching over these configurations can be quite important with larger \nvs ~domains as we can balance DP/TP/PP communications by re-distributing GPUs to take advantage of the faster domain.
    To get all possible configurations, similar to the parallelization, we decompose 
    $n_{\text{\nvs}} = n_{\text{\nvs1}} \times n_{\text{\nvs2}} \times n_{\text{\nvs p}} \times n_{\text{\nvs d}}$ into all possible factors. We ensure $n_{\text{NVSi}}$ divides $n_i$ and discard factors that do not.}
    \item \textbf{Additional SUMMA configurations}:
    \revision{
    For 2D TP {SUMMA}, we additionally include the number of blocked matrix multiplies $n_b$ (see \S\ref{sec:appendix_additional_methods} for details on the SUMMA algorithm and effect of the $n_b$ parameter) as a configuration to search. The possible values for $n_b$ is a function of the tensor dimensions and affects the {exposed} communication time for this strategy. 
    }
\end{enumerate} 

\subsection{Models and Systems Studied}
We study two classes of \emph{large} models---\gpt, an LLM foundation model, and long-sequence Vision Transformer \vit, a neural operator backbone for SciML foundation models. 
\gpt ~has $(l,e,h,d) = (2048,25600,160,128)$ and \vit ~has $(l,e,h,d) = (64800,12288,64,48)$. 
The first model is representative of foundation LLM pre-training with small sequence length $l$. The FLOP ratio of MLP to \SA ~is roughly $2\text{x}$. The \vit ~with large $l$ is representative of foundation models in science where the large $l$ arises from the necessity to process the entire spatial grid of a physical variable at a high resolution to ensure physical continuity and capture fine-scale phenomena \cite{willard2024analyzing, kurth2023fourcastnet, brenowitz2024practical}.  We base our hyperparameter choice on existing models \cite{nguyen2023climax,wang2024orbit}. For example, training on a popular weather forecasting dataset ERA5 \cite{hersbach2020era5}, the input resolution is a $720 \times 1440$ spatial grid giving rise to a 1M sequence length. We assume patch size $4$ (realistic spatial downsampling based on \cite{kurth2023fourcastnet,willard2024analyzing}) to drop $l$ to $64800$. At this scale, the \vit ~has a FLOP ratio of MLP to \SA ~as roughly $0.5\text{x}$, leading to large \SA ~operations, illustrating another extreme of foundation model pre-training. We note that long-context window LLMs may also represent this extreme, but they are typically only fine-tuned at this scale.
We assume that \gpt ~is pre-trained on 1T tokens  as planned for pre-training of LLMs for science \cite{auroragpt2,auroragpt3} and \vit ~is trained on 40 years of hourly data from ERA5 for 80 epochs \cite{willard2024analyzing}. In all experiments, we assume a global batch size of $4096$ samples.
For systems, we consider three generations of GPUs (\amp, \hop, \blackwell) to project trends as well as different \nvs ~domain sizes for each generation. We list the hardware (and network) characteristics of each system in  Tab. \ref{table:GPU_generation_params}. We also assume that the NVLink and \ib ~bandwidth increase proportionally across generations.

\section{Results}
\label{sec:results}

Our analysis centers on three questions:
% \noindent \textbf{(Q1)} 
\begin{enumerate}[label=(\textbf{Q\arabic*})]
\item What is the rationale behind an optimal configuration in the design space?
\item What are the optimal configurations and primary performance bottlenecks when training at scale and how do they change for different models? 
\item What is the effect of the GPU generation and \nvs ~domain size on overall performance? 
\end{enumerate}
% To simplify the vast parameter space, we fix the global batch size to 4096, restricting our analysis to the training of large datasets where sufficiently high global batch sizes can be afforded.

\begin{figure}
    \centering
    \includegraphics[width=\linewidth]{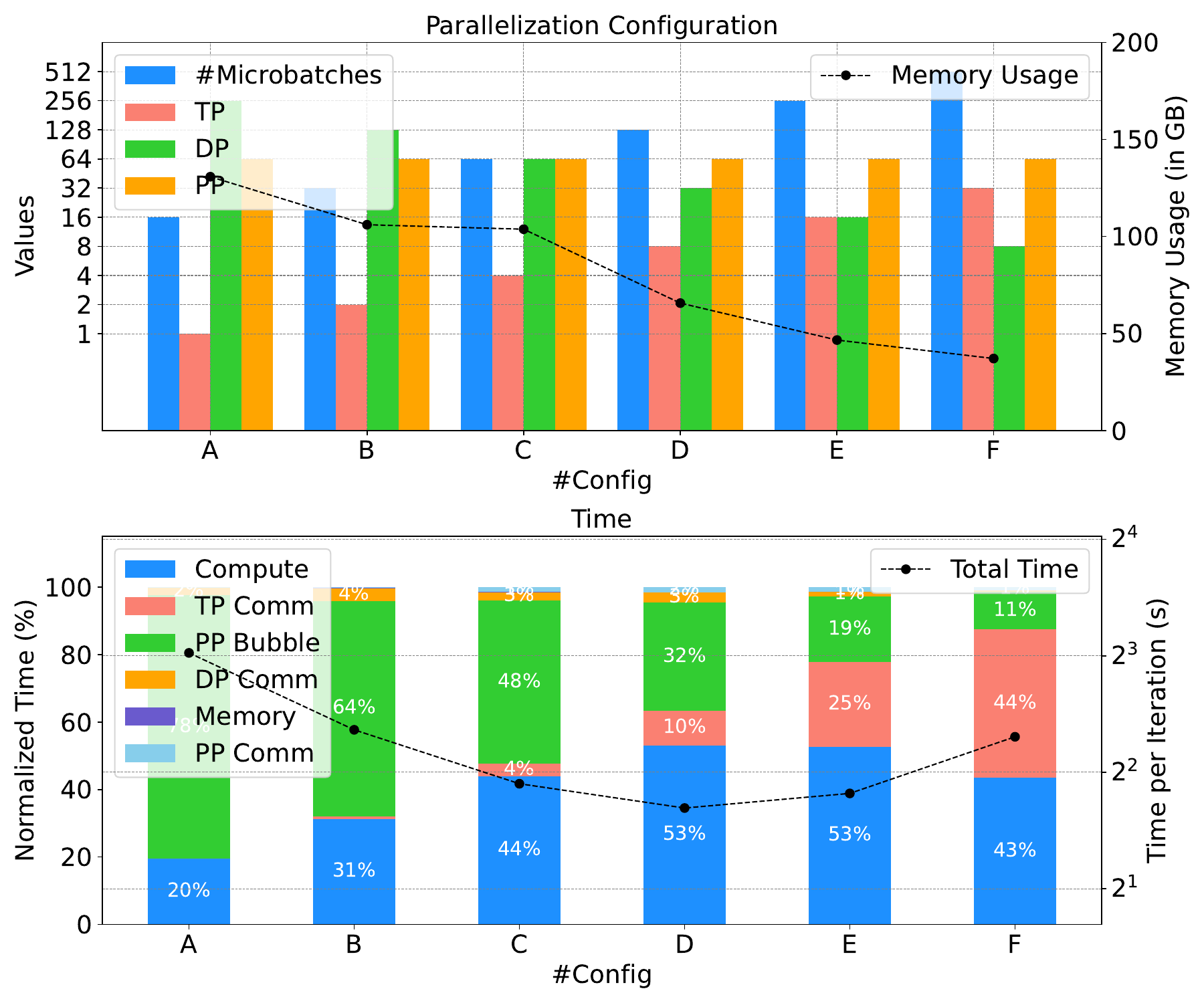}
    \caption{\emph{On $16384$ \blackwell  ~with \nvs ~domain size $n_{\text{\nvs}}=8$, microbatch size $1$ for \gpt ~with 1D TP: (top) Configurations (Config.) chosen with PP fixed to $n_p = 64$ and the others varying. Example, Config. D is $(m,n_t,n_d,n_p)=(128,8,32,64)$ and so on. (bottom) Time for each configuration broken down by the time spent on compute, memory accesses, various communications (DP, TP, PP) and PP bubbles. In config. D (TP $n_t=8$), we observe a local minimum in time per iteration.}}
    \label{fig:rationale}
\end{figure}

\begin{figure*}
    \centering
    \begin{subfigure}[b]{0.495\textwidth}
        \centering
        \includegraphics[width=\textwidth]{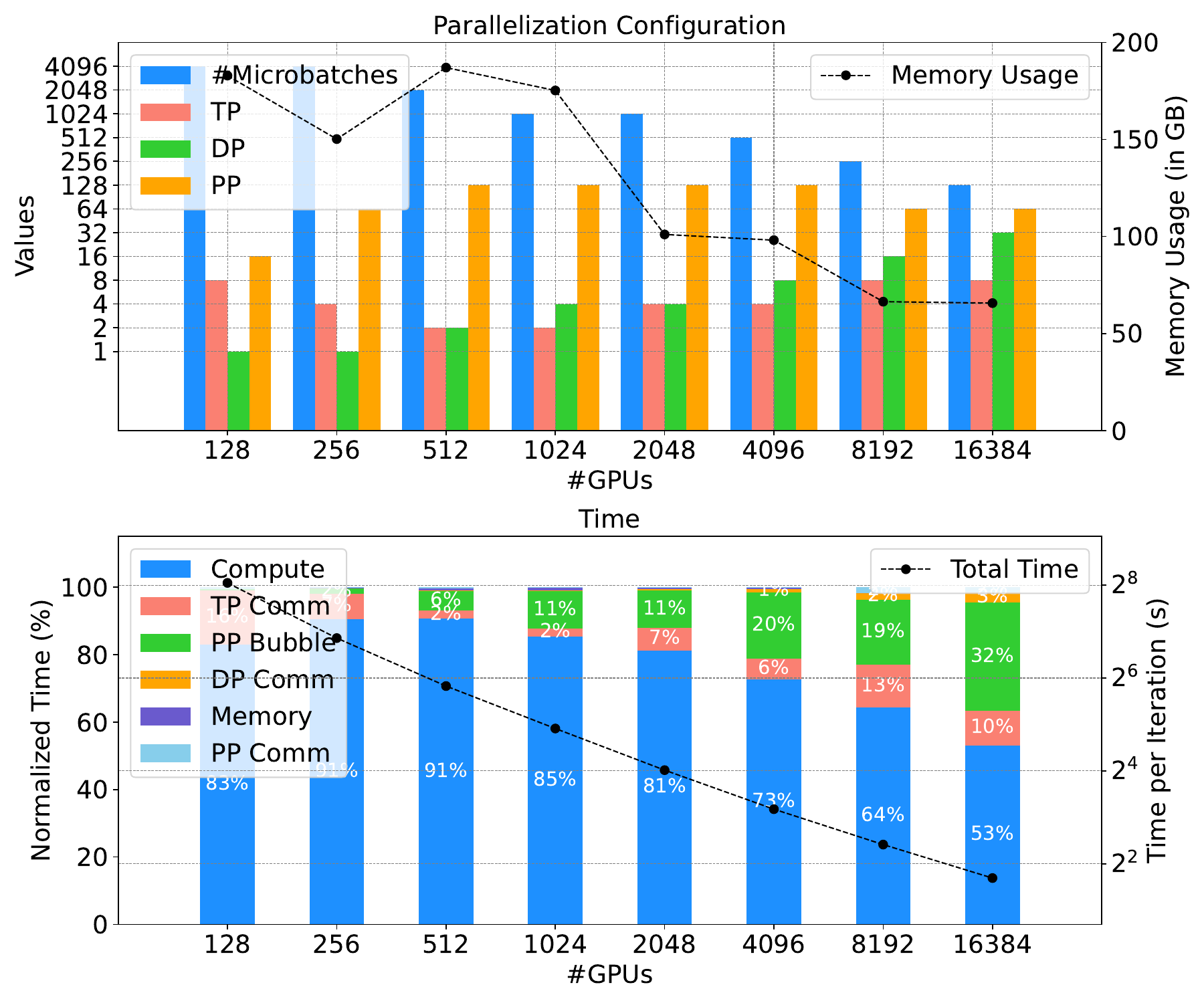}
        \caption{\textbf{\gpt ~with 1D TP}}
        \label{fig:gpt3-par}
    \end{subfigure}
    \hfill
    \begin{subfigure}[b]{0.495\textwidth}
        \centering
        \includegraphics[width=\textwidth]{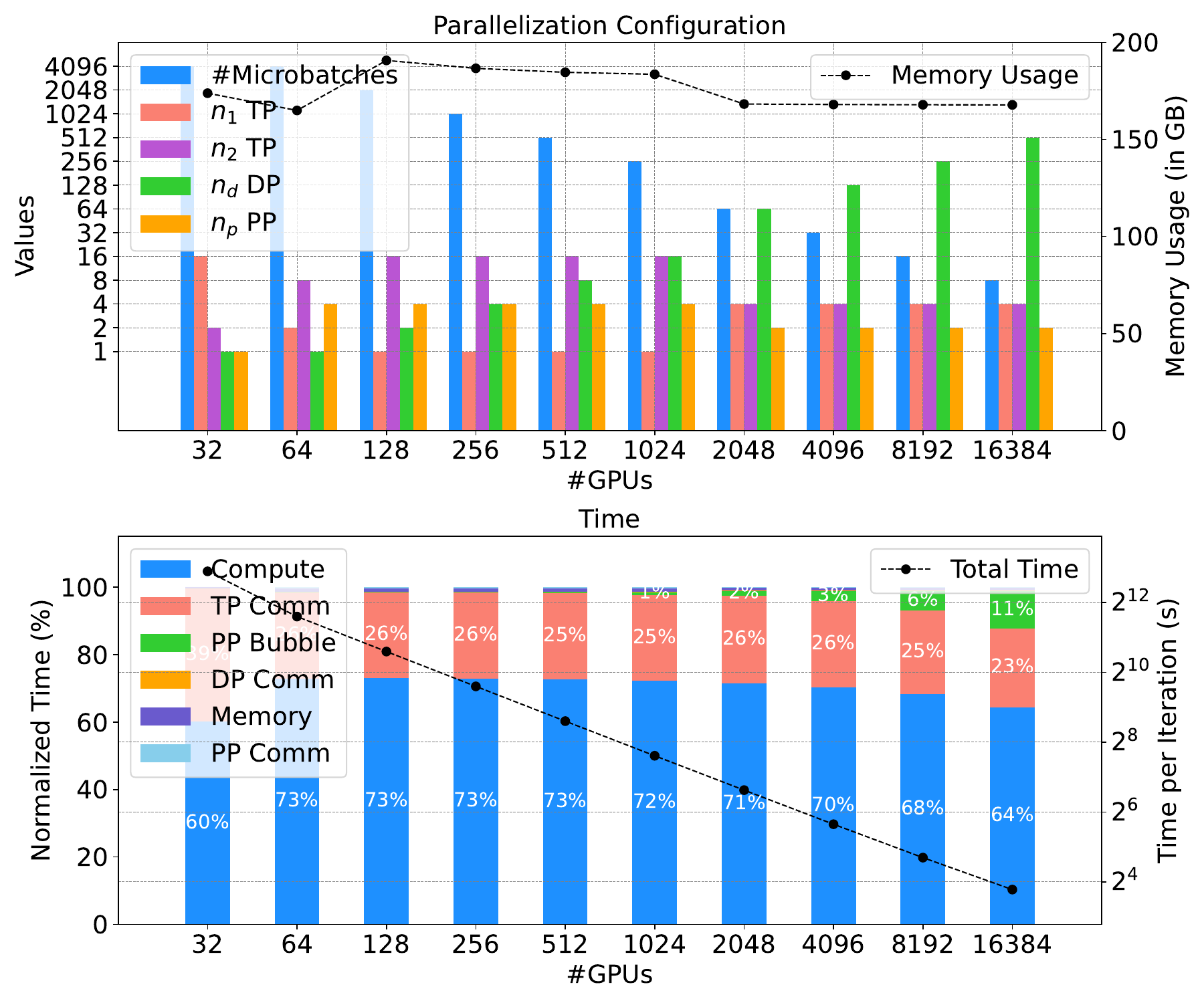}
        \caption{\textbf{\vit ~with 2D TP}}
        \label{fig:vit-par}
    \end{subfigure}
    \caption{\emph{On \blackwell ~with $n_{\text{\nvs}} = 8$: (top) Optimal parallelization strategy and \hbm ~memory consumed vs number of GPUs (bottom) Time vs number of GPUs broken down by the time. For both models, most time is spent in compute. For \gpt ~(left), 1D TP is sufficient to get good performance. PP bubbles start to dominate at scale, followed by TP and DP communication. \hbm ~memory used also drops at scale. For \vit ~(right), 2D-TP is necessary for fitting the model on \hbm ~and most of it is utilized, even at scale. TP communications are the main bottleneck followed by PP bubbles. 
    % Neither model shows significant memory access times or PP communication.
    }}
    \label{fig:par}
\end{figure*}

\revision{To provide a detailed view of optimal configurations and their resulting performance, we use plots similar to those in Fig. \ref{fig:rationale}. We first briefly explain the formatting of these plots for clarity. 
In the \textsc{Parallelization Configuration} plots (upper panels), 
the $y$ axis shows the parallelization configuration and the memory consumption on \hbm---we show the number of GPUs assigned to data parallel ($n_d$ DP), tensor parallel ($n_t$ TP or $n_1, n_2$ for 2D TP), pipeline parallel ($n_p$ PP) and the number of microbatches as vertical bars, and we show the total memory consumption in GB as a black dot.
On the $x$ axis, we either enumerate different configurations (example, Config. A, B, C, and so on) or enumerate number of GPUs used to train the model (when analyzing scaling performance as in Fig. \ref{fig:par}). Hence, the upper panels allow us to compare the details of the parallelization configurations and memory consumption for different settings.}

\revision{
In the \textsc{Time} plots (bottom panels), we show the same configurations as above, but with the vertical bars now displaying a time breakdown of each training step (in units of percentage of total iteration time). This allows one to visually compare across configurations and see how much time each configuration spends in a given stage (e.g. in pipeline bubbles, TP/DP/PP communication, compute, memory accesses). We also show the total time per iteration (black dots) to indicate which configuration is fastest. The lower panels allow us to inspect the bottlenecks in training and provide a sense of training efficiency.
}

\begin{figure}[t]
    \centering
    \begin{subfigure}[b]{0.5\textwidth}
        \centering
        \includegraphics[width=\textwidth]{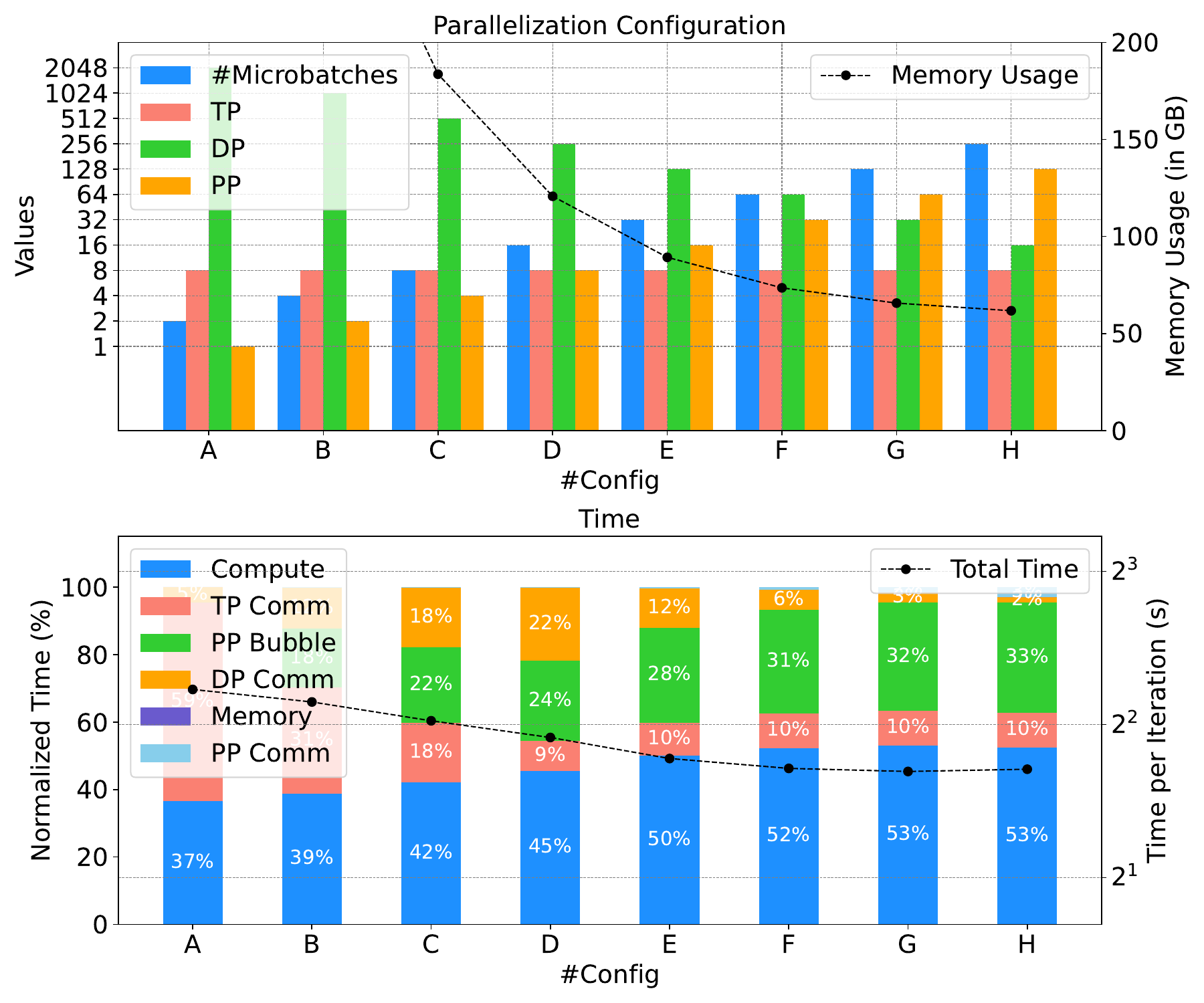}
        \caption{$n_\text{NVS} = 8$}
        \label{fig:gpt3-rat-nvs8}
    \end{subfigure}
    \hfill
    \begin{subfigure}[b]{0.5\textwidth}
        \centering
        \includegraphics[width=\textwidth]{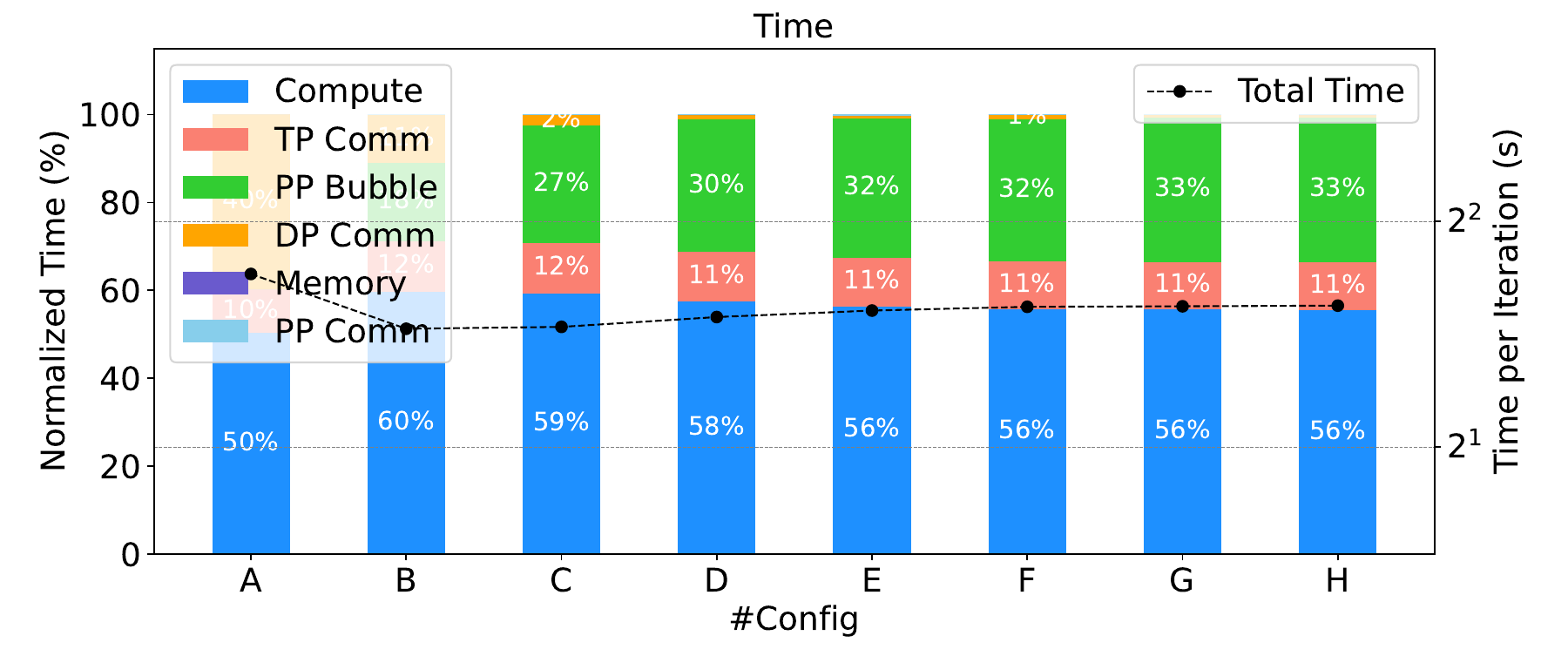}
        \caption{$n_\text{NVS} = 64$}
        \label{fig:gpt3-rat-nvs64}
    \end{subfigure}
    \caption{\emph{Following Fig, \ref{fig:rationale}, we change the following: we fix TP $n_t=8$ and vary PP and DP on two different \nvs ~domain sizes. (top) We observe a local minimum at PP $n_p=64$. (bottom) For larger \nvs, we observe that the local minimum  shifts to low PP $n_p=2$, with \nvs ~used to hide DP costs.}}
    \label{fig:rationale-nvs}
\end{figure}

\medskip \noindent \textbf{(Q1) Rationale behind optimal configurations. }
% We describe the different trade-offs in parallelization strategy in Fig. \ref{fig:rationale} assuming a fixed system (\blackwell ~and $n_{\text{\nvs}=8}$).  
\revision{It is challenging to visualize the entire design space to understand the rationale behind an optimal configuration. We make it easier to reason about design choices by fixing certain configurations and varying others. In subsequent analysis (\textbf{(Q2)}, \textbf{(Q3)}), we identify the optimal configuration by running the search \textbf{(S3)} over the entire design space. }

\revision{
For this analysis, we start with fixed total GPUs $n$ and global batch size $b$. Then, we vary any two parallelization parameters and keep the rest fixed. Since $n = n_1 \times n_2 \times n_d \times n_p$, as one configuration parameter increases, the other must decrease. Once we have chosen our parallelization configuration, we search over all possible GPU assignment configurations for each parallelization configuration to get the optimal assignment to the \nvs ~domain. This ensures that, for any parallelization configuration, the assignment to \nvs ~domain is optimal. We use the \gpt ~transformer for this analysis.
We make several observations.}
% \begin{enumerate}[label=(\textit{\textbf{\roman*}})]
    % \item 
    
    \medskip \noindent \textit{\textbf{(i)}}
    \revision{A simple observation is the apparent convex behavior of training time in the design space, especially w.r.t TP (tensor parallelism).}
    In Fig. \ref{fig:rationale}, assuming a fixed system (\blackwell ~and $n_{\text{\nvs}}=8$) and 1D TP, we fix PP $n_p=64$ and vary both TP and DP (with the total GPUs fixed at $n = 16384$ and microbatch size 1), to get training time as a function of varying TP/DP. We can immediately see the convexity of time that arises due to dominant costs from increased TP and DP. \revision{Note that as DP decreases, since global batch size $b$ and microbatch size $b_m$ are fixed, the number of microbatches ($m$) automatically increases}. We see that greater TP drops the \hbm ~usage but exhibits high communication costs (due to more microbatches). With large DP, number of microbatches reduces leading to large pipeline bubbles. The DP and PP communications are sufficiently small and hidden here. Hence, there is a local minimum around $n_t=8,n_d=32,m=128$ with about 40G \hbm ~utilization.
    % \item 

    \medskip \noindent \textit{\textbf{(ii)}}
    \revision{The dual-bandwidth domain introduces \emph{subtle non-convexities} in the training time within the design space. }
    We show this by fixing TP at $n_t=8$ (for 1D TP), and varying PP and DP in Fig. \ref{fig:rationale-nvs}. As DP increases, the DP communications follow a non-convex pattern, increasing to a peak transition point and then declining. This is due to the dual bandwidth domain where the chosen $n_\text{\nvs d}$ (number of DP GPUs on \nvs) starts to increase at the transition point---both TP and DP begin to utilize the fast bandwidth, decreasing the DP communication time. With maximum DP, all 8 \nvs ~GPUs are utilized for DP. The local minimum is still the same.
% \end{enumerate}
% In Fig. \ref{fig:rationale}, assuming a fixed system (\blackwell ~and $n_{\text{\nvs}=8}$), we fix PP $n_p=64$ and vary both TP and DP (with the total GPUs fixed at $16384$ and microbatch size 1) to demonstrate a known convexity w.r.t TP. 
% We see that greater TP drops the \hbm ~usage but exhibits high communication costs (due to more microbatches). With large DP, number of microbatches reduces leading to large pipeline bubbles. The DP and PP communications are sufficiently small and hidden here. Hence, there is a local minimum around $n_t=8,n_d=32,m=128$ with about 40G \hbm ~utilization.
% However, fixing TP at $n_t=8$, and varying PP and DP in Fig. \ref{fig:rationale-nvs}, the convexity becomes more subtle. As DP increases, the DP communications follow a non-convex pattern, increasing to a peak transition point and then declining.
% This is the effect of the dual bandwidth domain where the chosen $n_\text{\nvs d}$ (number of DP GPUs on \nvs) starts to increase at the transition point---both TP and DP begin to utilize the fast bandwidth. With maximum DP, all 8 \nvs ~GPUs are utilized for DP.

\medskip \noindent \textit{\textbf{(iii)}}
\revision{Larger \nvs ~domains can exploit the above by favoring alternate parallelization strategies.}
To show this, we repeat the the above experiment for $n_{\text{\nvs}}=64$ in Fig. \ref{fig:gpt3-rat-nvs64}. We observe that the optimal configuration shifts to heavily decreased PP with the entire \nvs ~domain used for DP and TP. Hence, the larger \nvs ~domain has favored increased data parallelism with minimal pipelining, at the cost of increased HBM utilization. Note that while $n_p = 1$ is fastest, it is infeasible on a \blackwell ~GPU due to high HBM capacity required.
% Smaller PP than this increases the exposed DP communications due to larger number of layers per GPU.
% At the same time, \hbm ~capacity restricts the feasible configuration and we observe that larger \nvs ~domains favor lesser PP at the expense of larger HBM utilization.

% Hence, while the iteration time generally shows convex behaviours, there are \emph{subtle non-convexities} due to the dual bandwidth domains that are emphasized as the \nvs ~domain sizes increase. 
\begin{figure}
    \centering
    \begin{subfigure}[b]{0.5\textwidth}
        \centering
        \includegraphics[width=\textwidth]{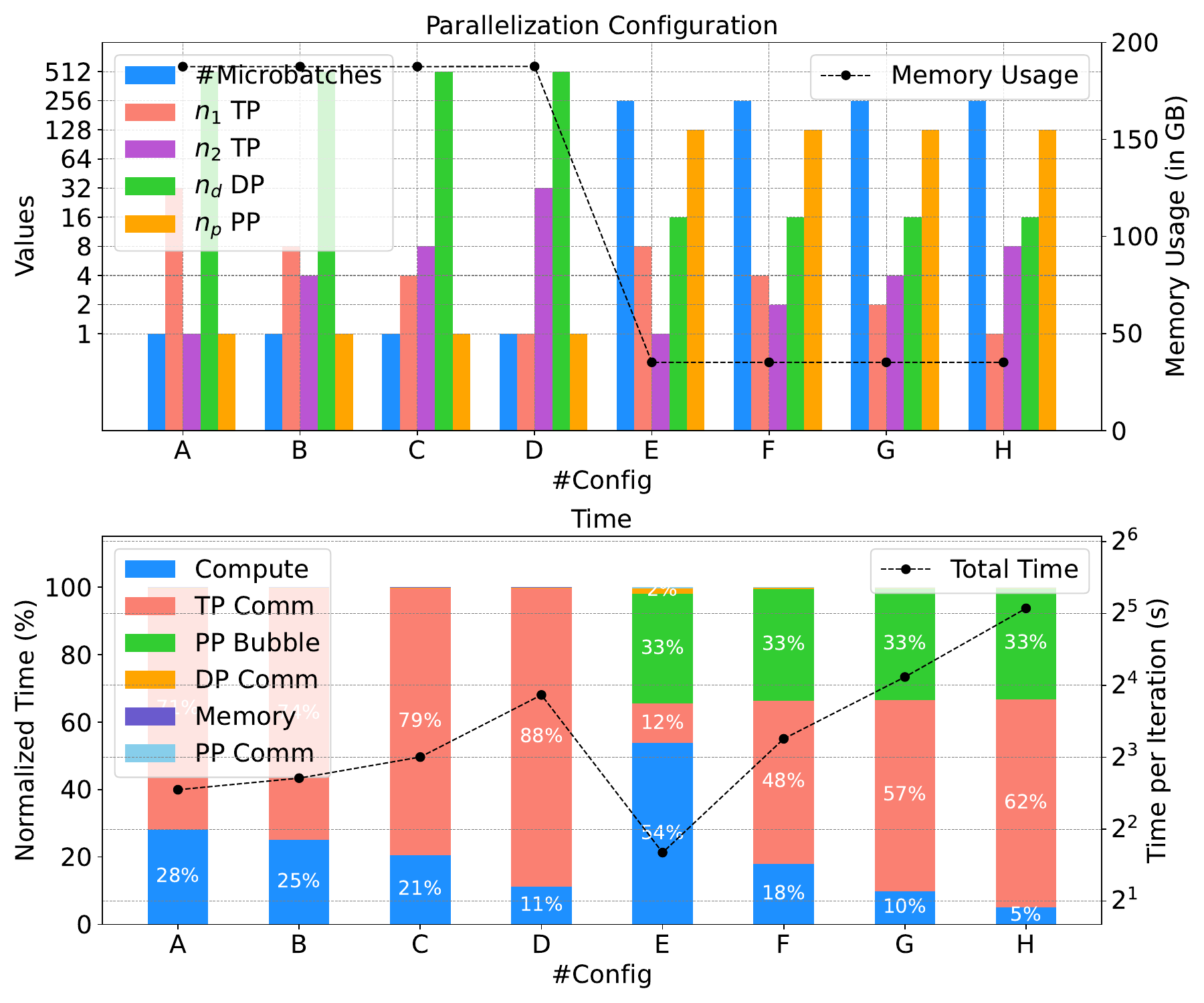}
        \caption{$n_\text{NVS} = 8$}
        \label{fig:gpt3-2d-rat-nvs8}
    \end{subfigure}
    \hfill
    \begin{subfigure}[b]{0.5\textwidth}
        \centering
        \includegraphics[width=\textwidth]{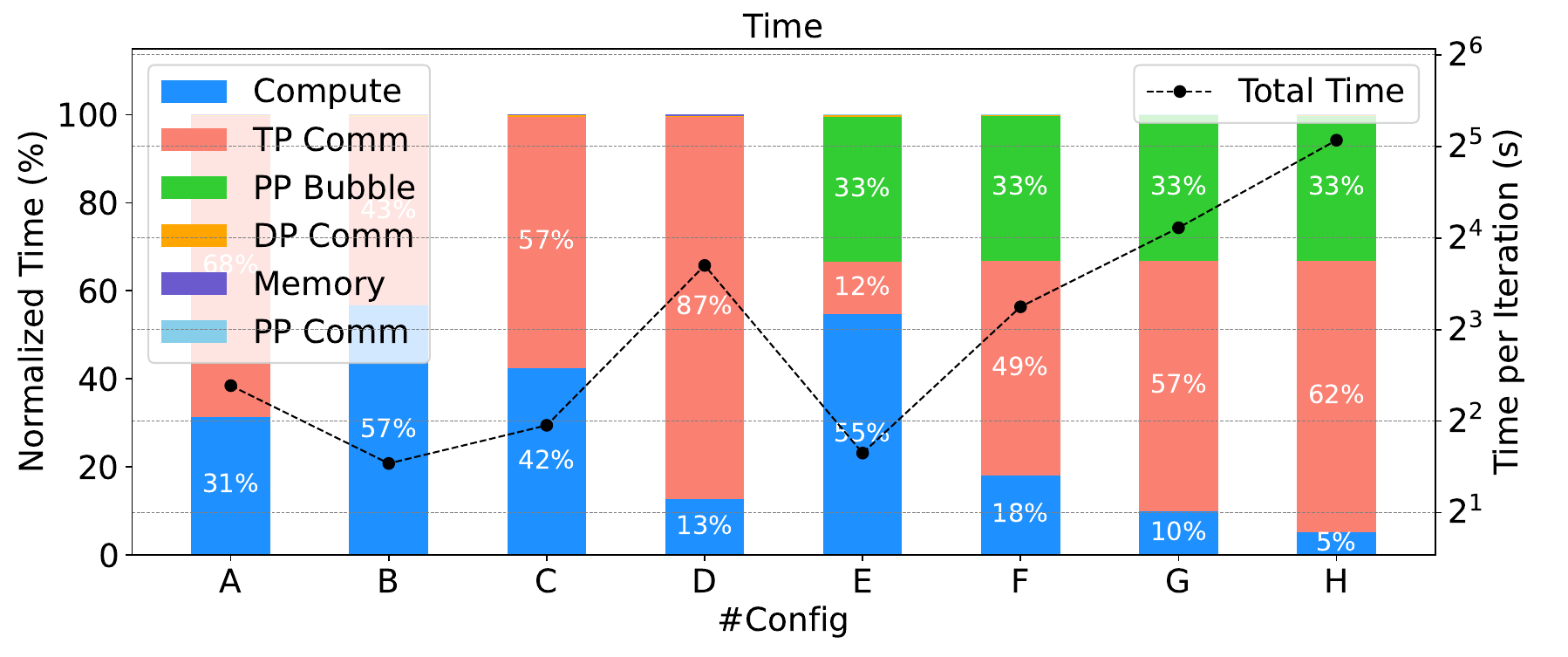}
        \caption{$n_\text{NVS} = 64$}
        \label{fig:gpt3-2d-rat-nvs64}
    \end{subfigure}
    \caption{\emph{For \gpt ~with 2D TP SUMMA, we fix TP $(n_t,n_p)=(32,1)$ and vary $n_1, n_2$ to get the the first five configurations and then switch to $(n_t,n_p)=(8,128)$ and repeat the same for the rest. We show two \nvs ~domain sizes. With larger \nvs, large DP (low PP) is preferred.}}
    \label{fig:rationale-nvs-2d}
\end{figure}

\medskip \noindent \textit{\textbf{(iv)}}
\revision{Higher dimensional 2D TP versions can show similar behaviors with the design-space being more complex due to additional TP dimensions.}
We sweep over two sets of configurations in Fig. \ref{fig:rationale-nvs-2d} to show this in 2D TP {SUMMA}. We focus on the two extremes we observed in the 1D case (high PP and high DP). We first set high DP by fixing $n_p = 1$, one microbatch $m = 1$, and choosing a large enough TP $n_t = 32$. \revision{We note that the microbatch size is now 8 and the memory efficiency of 2D TP SUMMA enables us to fit the model on HBM; 1D TP would need far greater TP for this microbatch size and $n_p = 1$ due to shared activation memory, leading to unmanageable communication costs}. We then only vary the \emph{relative} TP allocation into $n_1$ and $n_2$. Next, we switch to low DP by setting $n_p = 128$ and repeat with $n_t = 8$. We retain large $m$ here to manage the pipeline bubbles (since $n_p > 1$).
We observe that in both high and low DP configurations the fastest configuration involves only 1D TP with $n_2 = 1$. This is because adding the second dimension increases the communication volume significantly in SUMMA (as noted in \S\ref{sec:methods}) and, while more GPUs can decrease this (since SUMMA communication volumes scale with number of GPUs), the slow \ib ~bandwidth creates bottlenecks. In Fig. \ref{fig:gpt3-2d-rat-nvs8}, $(n_1,n_2,n_p) = (8,1,128)$ is fastest. Increasing the \nvs ~domain size to $64$, in Fig. \ref{fig:gpt3-2d-rat-nvs64}, favors high DP (similar to 1D TP) with $(n_1,n_2,n_p) = (8,4,1)$ being fastest. The fast bandwidth helps manage and balance the large TP costs from both dimensions with effectively no PP. \revision{ Note that, here, the DP communications are effectively hidden behind the increased compute of a large microbatch size.} 2D TP is also similar but shows higher memory pressure from shared weights and activations (see Fig. \ref{fig:2d-tp-rationale} for details).

% \medskip \noindent \textit{\textbf{(v)}}
% \revision{The large \nvs ~domain emphasizes more DP in 2D TP versions as well (similar to 1D) with more balanced TP dimensions.}
% The two attractors above are comparable in Fig. \ref{fig:gpt3-2d-rat-nvs64} where $(n_1,n_2,n_p) = (8,4,1)$ now contends with the other configuration because the fast bandwidth has made TP communications manageable in 2D TP SUMMA. This is comparable to the 1D TP case where larger NVS domains favor smaller PP. 
% 2D TP also shows similar behaviors. 
% However, despite good performance, the second local minimum is no longer a feasible configuration for 2D TP (see Fig. \ref{fig:2d-tp-rationale}) as there is a lot of memory pressure from shared weights and activation maps---with SUMMA, the memory utilization is more efficient. 

% Hence, the parallelization strategy, GPU assignment to \nvs ~domain (and its size) can significantly alter the design space leading to different rationales for the optimal strategy. With \vit, we see similar behavior but with larger memory pressure (see Fig. \ref{fig:2d-tp-rationale} and associated discussions). 
% We discuss model-specific changes next.

\medskip \noindent \textbf{(Q2) Optimal parallelization strategy as a function of transformer type. }We show the \emph{optimal} parallelization strategy from the performance model as a function of number of GPUs $n$ in Fig. \ref{fig:par} on \blackwell ~with $n_\text{\nvs}=8$. \revision{To get the optimal configuration (note that this includes the parallelization configurations as well as assignment of GPUs to \nvs ~domains), we run our search (\textbf{S3}) for each $n$, independently. In these plots, we also note the difference that the $x$ axis now enumerates an increasing number of GPUs on which the models are parallelized, and for each scale ($n$) we show the optimal configuration and time breakdown in the same format as previous figures.
These plots show the strong scaling behavior of the transformer and highlights how parallelization configurations and training time bottlenecks change with the use of more GPUs for training. 
We make the following observations.
} 
% We also show the time broken down into various components.
% \revision{We first focus on the \gpt ~transformer and observe the following:}

\medskip \noindent \textit{\textbf{(i)}}
\revision{The optimal parallelization configurations for \gpt ~are re-balanced as $n$ changes due to different bottlenecks arising at different scales.}
We show this in Fig. \ref{fig:gpt3-par} assuming a \blackwell ~system with $n_{\text{\nvs}} = 8$. At small GPU scales (128, 256), the performance model opted to use increased TP to fit the model, leading to large TP communication. At larger scales, TP reduces and then increases monotonically. PP increases to the maximum value (depth $d$) until 4096 GPUs. 
At this point, DP has increased to the point that the number of microbatches (which reduce with more DP) are not large enough to hide the pipeline bubbles, seen in the increasing pipeline bubble fractions, and PP starts to reduce. Smaller PP exposes more TP communication and hence the optimal configuration carefully re-balances each of these at different scales to manage the bottlenecks.
Finally, DP and TP communications are insignificant at small scale, but at large scale they slowly get exposed. 

\medskip \noindent \textit{\textbf{(ii)}}
Each GPU scale shows a different optimal assignment of GPUs onto the \nvs ~domain for \gpt. For example, at 2048 GPUs, the performance model opted $(n_{\text{\nvs1}},n_{\text{\nvs p}}) = (4,2)$ (note that \nvs ~size is $n_{\text{\nvs}} = 8)$, at 1024 GPUs we see $(n_{\text{\nvs1}},n_{\text{\nvs d}},n_{\text{\nvs p}}) = (2,2,2)$  and the full \nvs ~domain is used for TP at 8192 GPUs and beyond. Hence, the fast bandwidth is exploited differently across various $n$ for each parallelization strategy, to better hide their respective dominating costs.

\medskip \noindent \textit{\textbf{(iii)}}
HBM capacity utilization is high only at small-to-moderate GPU scales and drops at larger scale for \gpt. We also note that memory access time is insignificant at all scales for this transformer with compute being the dominant cost, followed by pipeline bubbles, TP communication and small DP/PP communications at large scale.
% We observe that the \hbm ~capacity is not fully utilized at large scales and the increased parallelism is favored for faster compute.
2D TP versions show similar behavior with re-balanced configurations, HBM utilization, and dominant bottlenecks at different scales (see Fig. \ref{fig:gpt-all} as an example).
% As we switch to 2D TP, the optimal configuration is still 1D TP due to increased communication time across \ib ~(as noted in \textbf{Q1}). With larger \nvs, we see greatly reduced PP (see Fig. \ref{fig:gpt-all}) in 2D versions as well.

\medskip \noindent \textit{\textbf{(iv)}}
In \vit, sequence length $l = 64\text{K}$ renders 1D TP \emph{infeasible} on all GPUs due to extremely large activation map memory, with 2D TP necessary and dominating the optimal configurations. We show the \vit ~2D TP in  Fig. \ref{fig:vit-par}. Due to large $l$, large TP is necessary with $n_1=4$ and $n_2=4$ as the most common strategy at scale. PP is low and TP communications are the only bottlenecks. While PP helps reduce weights memory, it increases activation memory as noted in \S\ref{sec:methods} as the 1F1B schedule stores $n_p$ microbatches of activation maps. Only TP helps in this model setting at the cost of high communication time---the full \nvs ~domain is used only for TP. \hbm ~capacity is also highly utilized. 2D TP {SUMMA} exhibits similar behaviors with a smaller throughput (due to increased communication) but is more memory efficient. 

% Hence, the optimal configurations are sensitive to not only the scale of GPUs used but also the transformer model type, exposing different parallelization bottlenecks. 
% Next, we look at this sensitivity as the GPU generation changes (along with the network and \nvs ~domain size).

\begin{figure}
    \centering
    \begin{subfigure}[b]{0.5\textwidth}
        \centering
        \includegraphics[width=\textwidth]{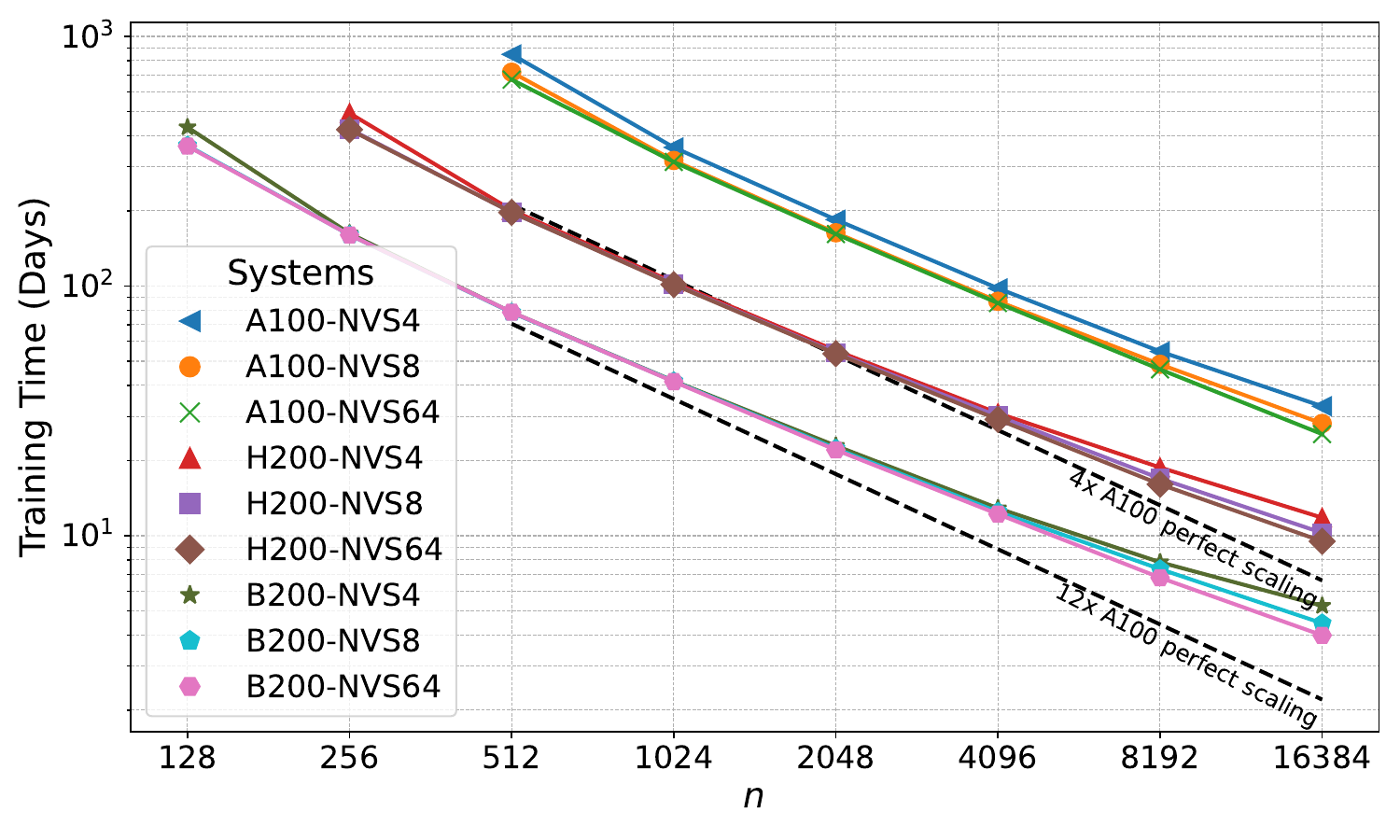}
        \caption{\textbf{GPT3-1T with 1D TP}}
        \label{fig:gpt3-time}
    \end{subfigure}
    \hfill
    \begin{subfigure}[b]{0.5\textwidth}
        \centering
        \includegraphics[width=\textwidth]{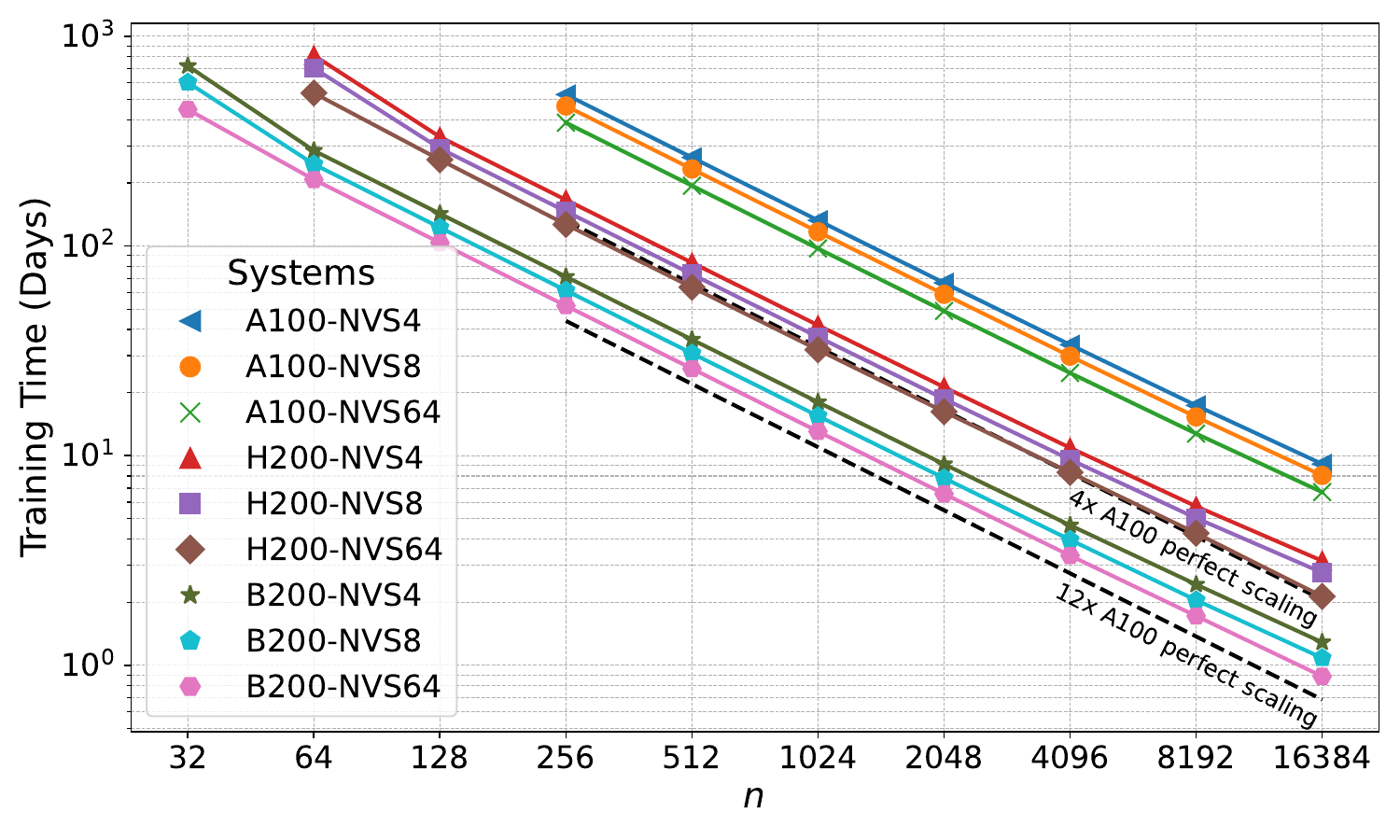}
        \caption{\textbf{ViT with 2D TP}}
        \label{fig:vit-time}
    \end{subfigure}
    \caption{{\emph{Training time in days vs number of GPUs ($n$) for different GPUs (\amp, \hop, \blackwell). We also assume different \nvs ~domain sizes for each generation---\nvs4 indicates $n_\text{\nvs}=4$ and so on (top) Training time on 1T tokens for \gpt ~using 1D TP. We see a consistent improvement across GPU generations. Further, smaller number of future GPUs are capable of training \gpt ~due to larger \hbm. \nvs ~effects are seen at the smallest scale (due to increased TP) and at the largest scales. (bottom) Training time on 40 years of ERA5 with a \vit ~using 2D TP. While GPU generations provide similar advantages, \nvs ~domain size effects are seen throughout due to increased TP necessary for this model.}}}
    \label{fig:overall-time}
\end{figure}

\medskip \noindent \textbf{(Q3) Overall Performance, GPU Generation, and NVS Domain Size. }We show the training time vs number of GPUs for the two models for different GPU generations (\amp, \hop, \blackwell) as well as different \nvs ~domain sizes (4, 8, and 64) in Fig. \ref{fig:overall-time}. We make the following observations:

\medskip \noindent \textit{\textbf{(i)}}
The \gpt ~model benefits significantly with higher GPU generation.
We show this in Fig. \ref{fig:gpt3-time}, where the training times are $\mathcal{O}$(30) days on $16K$ \amp ~GPUs and it drastically drops to $\mathcal{O}$(3-5) days on the \blackwell ~GPU, owing to increased tensor core performance and network bandwidths. We also show this in Figs. \ref{fig:flops-hbm} and \ref{fig:hbm} where we sweep over different hardware characteristics and highlight the reduced sensitivity to capacity/HBM bandwidth and also show that configurations with small memory bandwidth and high capacity can also show good performance, highlighting the value of alternate memory technologies. Pre-training this model would require $>10K$ GPUs to keep training reasonable at $\mathcal{O}$(days), even for the \blackwell, emphasizing the importance of understanding performance bottlenecks at scale. 

\medskip \noindent \textit{\textbf{(ii)}}
The \nvs ~domain effects are only seen at large scale (and increasing with scale) for the \gpt ~model, where the larger domains increase the scalability of the model.
% This is because TP is always chosen within \nvs, irrespective of its size, with PP bubble inefficiencies favored.
As noted in \textbf{Q1}, this is because larger \nvs ~domain allows large DP (at the cost of higher memory usage) by hiding the DP communication costs which, in turn, decreases pipeline bubbles (that dominate the bottlenecks at scale) thereby improving scalability. Hence, with large \nvs ~as well, TP remains small with the \nvs ~domain utilized for DP to scale better (see Fig. \ref{fig:gpt-all}).
% We also note that larger \nvs ~does not mean larger TP is better. In fact, TP is always chosen within \nvs, irrespective of its size, as PP is favored more.
% Even with 2D versions of TP, the TP does not increase to much larger scales with more GPUs (see Fig. \ref{fig:gpt-all}). Instead PP/DP are still employed at scale for an optimal configuration.
At the small scale, the \nvs ~domain again plays a role due to heavy parallelism to fit the model on available GPUs. At moderate scale, the effects are milder---fine-tuning jobs at this scale may not see a lot of benefit from larger \nvs ~domains. 
\revision{Similarly, the \nvs ~domain helps more for the \amp ~GPUs due to its limited capacity, leading to increased TP}.

\medskip \noindent \textit{\textbf{(iii)}}
\revision{While 1D TP is performant for \gpt, both 2D TP versions can help further reduce training time (see Fig. \ref{fig:speedup} for relative speedups). We observe that 2D TP SUMMA can particularly help in the resource constrained regime---smaller number of GPUs, small capacity (as in \amp), and small \nvs ~domain size. The speedups reduce as the GPU generation increases.  
Similar to 1D, as the \nvs ~domain becomes larger, the 2D TP versions can also utilize it in favor of increased DP (see Fig. \ref{fig:gpt-all}), increasing the scalability of the model.}
% Even with 2D versions of TP, the TP does not increase to much larger scales with more GPUs (see Fig. \ref{fig:gpt-all}). Instead PP/DP are still employed at scale for an optimal configuration.
% As noted in Fig. \ref{fig:gpt3-par} (and Fig. \ref{fig:flops-hbm}), at large scale the \hbm ~capacity also plays less of a role. 
% Hence, the \gpt ~class models exhibit different behaviors based on scale: at smaller scales (fine-tuning workloads, inference workloads), \hbm capacity is important while NVS domain size is less critical, whereas at larger scales, larger NVS domains can become quite critical, and HBM capacity becomes less significant. 

\medskip \noindent \textit{\textbf{(iv)}}
The \vit ~also sees significant improvements with GPU generation. However, the 2D versions of TP are the only viable strategies, with 2D TP being optimal. Both \hbm ~capacity as well as \nvs ~domain size effects show more uniform importance across scales (also see Fig. \ref{fig:vit-flops-hbm}). This is due to the increased pressure placed on TP to fit long sequences. For this model, since $n_t=16$ is necessary any \nvs ~domain size less than this will show performance drops. Hence, larger \nvs ~domains add consistent value until $n_t$ GPUs are within the same domain, after which there are diminishing returns.
% After this, there are diminishing returns since PP is not used and DP communications are sufficiently hidden behind compute at all scales. 
We show that the \vit ~also shows alternate large capacity/low memory bandwidth configurations (alternate memory to \hbm) as viable options in Fig. \ref{fig:vit-hbm}.

\medskip \noindent \textbf{Empirical Validation. }
% An in-depth empirical validation is out of scope of this paper. However, 
We verify that the performance model produces reasonable outputs for the different model types and parallelization strategies at scale. We consider moderate scale tests on 512 GPUs with global batch size 1024 on the Perlmutter \cite{perlmutter} supercomputer at NERSC. Perlmutter has 4 \amp ~GPUs per node (all-to-all connected via NVLink). We validate the performance model with a 175B parameter \textsc{GPT3} and a 32K \vit ~model using \textsc{Megatron-LM} \cite{megatroncolde}.
% where different parallelization configurations are set.  
% The GPT model is implemented in the the Megatron-LM \cite{megatron} codebase. Training is done in a 1F1B non-interleaved pipeline schedule wth optimal distributed optimizer settings.  Efficient implementation of the GPT-like model leverages TransformerEngine and FlashAttention-2. 
We validate the optimal configuration from the performance model as well as few sub-optimal configurations.
For \textsc{GPT3}, we observe that the optimal configuration $(n_t,n_p,n_d,b_m) = (4,16,8,1)$
shows a 11\% error in iteration time.
We test 4 other sub-optimal configurations (with different relative TP/PP/DP) and they show 4--15\% errors.
% We also change the relative assignment of $n_t, n_p, n_d$ (increasing one and decreasing others) and observe around 4--15\% errors. While the latter configurations are sub-optimal, they show that the model is able to reasonably predict performance drops if such configurations are chosen. 
% For 2D TP, we consider a 32K sequence length version of the \vit ~on 512 GPUs (1024 batch size).  For a near-optimal configuration $(n_1, n_2, n_p, n_d, b_m) = (2, 4, 4, 16, 1)$ (the optimal configuration overflowed \hbm ~due to extra scaffolding memory in PyTorch; we picked the next configuration), the error is about 2\%. However, sweeping through different assignments of sub-optimal configurations, we observe the errors to ranging from 11--26\%.
For the \vit, a near-optimal configuration $(n_1, n_2, n_p, n_d, b_m) = (2, 4, 4, 16, 1)$ (the optimal configuration overflowed \hbm ~due to extra scaffolding memory in PyTorch; we picked the next configuration), the error is about 2\%. For sub-optimal configurations, with different TP/PP/DP, the error ranges from 11--26\%. 
% \revision{Some additional optimizations, such as overlapping TP communications, could likely contribute to this error range.}
We observe performance trends between observed and predicted iteration times are consistent (larger observed times seen with larger predicted times)
% This may be due to overlapping of some $n_2$ communications with compute in Megatron-LM context parallel implementations through Ring Attention variants that we do not model here
% This is likely due to larger predicted $n_2$ communications where we do not model overlap between this communication and \LA ~compute, which \textsc{Megatron-LM} implements through newer ring attention variants.
%
and the low errors are encouraging. 

\section{Summary and Discussion}
\label{sec:summary}
 
We have described the core components of an analytical performance model and its sensitivity to the transformer architecture, system features, and parallelization strategy.
We observe the following high-level characteristics:
% \begin{enumerate}[label=(\textit{\textbf{\roman*}})]
% \item 
\textit{\textbf{(i)}}
Placement of GPU groups within \nvs ~domain can significantly affect the performance, motivating software codebases to be flexible in not just the parallel configuration, but also the GPUs used for the configuration\footnote{\textsc{Megatron-LM} already provides some flexibility to the user in this aspect.};
% \item 
\textit{\textbf{(ii)}}
\gpt ~class models can benefit greatly with larger \nvs ~domains in pre-training scales ($>10K$ GPUs), where $\mathcal{O}$(days) improvements can be significant;
% \item 
\textit{\textbf{(iii)}}
Smaller (and moderate) scales of fine-tuning demand more from \hbm ~and less from \nvs ~($\mathcal{O}$(hours) improvements), suggesting different system design choices based on the training regime of large models. \revision{SUMMA variants of TP can particularly help at small scales and in resource-constrained regimes (small capacity, small \nvs)};
% \item 
% \item 
\textit{\textbf{(iv)}}
\vit ~class models demonstrate a contrasting extreme with more dependence on \nvs, \hbm, and higher-dimensional parallelism (with 2D TP variants). SUMMA variants can reduce some memory pressure, but heavy dependence on expensive system features may motivate algorithmic advances for reduced sequence length training;
% \item 
\textit{\textbf{(v)}}
Both models benefit most from FP16 tensor core performance and network bandwidths, but other memory technologies (lower bandwidths/more capacity like \textsc{LPDDR}) can be viable alternatives which may help alleviate the heavy dependence on the \nvs ~for \vit ~class models.
% \end{enumerate}
 % However, the \hbm ~is less important at this scale and alternate memory technologies like \textsc{LPDDR} could be considered. 
 % ---a supercomputer primarily serving fine-tuning needs may not benefit from large \nvs ~domains. 
 % We describe few limitations to our analysis in \S\ref{sec:appendix_limitations} around other possible optimizations in parallelization that we do not consider here.

 \medskip \noindent \textbf{Limitations. }
 \revision{While several components are modeled in the performance model, there are some optimizations we do not consider. We do not consider interleaved pipeline schedules \cite{Narayanan:2021} that can drop bubble time further. There are more lower-level opportunities for TP communications to be overlapped with compute to drop communication time. 
Weights (and gradients) can also be partitioned using DP at the cost of higher communication introducing another trade-off. We also do not explore offloading to the CPU (in addition to \hbm) which may be very useful for large sequences.}
 
 \medskip \noindent \textbf{Outlook. }
 % to understand better the interaction between architecture, parallelism, and system across a more diverse domain range outside of NLP.
% While several components are modeled here, there are some limitations. We do not consider interleaved pipeline schedules that can drop bubble time more. Further, certain TP communications can be overlapped with compute to drop communication time. Other optimizations such as partitioning the weights using DP GPUs are also possible. We focus our future work in including more SOTA optimizations and a diverse range of architectures, not restricted to transformers, to understand the interaction between architecture, parallelism, and system.
Overall, this work emphasizes the need to closely explore the complex interplay between architecture type beyond LLMs, parallelization strategies beyond 3D to 4D (and higher) versions, and the HPC system (interconnects and accelerator features), and demonstrates a path towards a tightly coupled modeling of all these elements.
 \revision{We focus our future work in including more optimizations, detailed analyses of 2D TP versions, and other architecture types such as linear (or windowed) attention versions of the \vit, spectral transformer models, convolutional models, graph neural networks, amongst other scientific AI foundation models.}

\section*{Acknowledgements}
This research used resources of the National Energy Research Scientific Computing Center (NERSC), a Department of Energy Office of Science User Facility using NERSC award ASCR-ERCAP0027797. 
The views and opinions of authors expressed herein do not necessarily state or reflect the position or the policy of our sponsors and no official
endorsement should be inferred. 
SS would like to thank Josh Romero from NVIDIA and Suvinay Subramanian from Google for valuable discussions on different aspects of performance modeling.

\clearpage
\bibliographystyle{IEEEtran}
\bibliography{references}
\appendix
\renewcommand\thefigure{\thesection\arabic{figure}}    
\setcounter{figure}{0}    
\setcounter{table}{0}
\renewcommand{\thetable}{A\arabic{table}}
% \subsection{Limitations}
% \label{sec:appendix_limitations}
% While several components are modeled in the performance model, there are other optimizations we do not consider. We do not consider interleaved pipeline schedules \cite{Narayanan:2021} that can drop bubble time or more complex schedules that can theoretically reduce bubble time to very small fractions. Further, certain TP communications (example, in \LA) can be overlapped with compute to drop communication time. 
% For large models, weights can also be partitioned using DP at the cost of higher communication introducing another trade-off. We also do not explore offloading to the CPU (in addition to \hbm) which may be very useful for large sequences. 
% Since our performance model is analytic, our analysis is inherently first order.
% We focus our future work in including more SOTA optimizations and architecture types such as linear attention versions of the \vit, convolutional models, graph neural networks to understand more the interaction between architecture, parallelism, and system across a more diverse range that is present in domains outside of NLP.

\subsection{Methods: Additional Details}
\label{sec:appendix_additional_methods}
\revision{We outline the abbreviations used in this paper in Tab. \ref{table:abbrev}.}

\begin{table}[htbp]
  \centering
  \rowcolors{2}{gray!25}{white} % Alternating row colors
	\begin{tabular}{l | l } 
		% \hline
		Abbreviation & Description\\ 
		\hline
		\nvs & NVSwitch \\
            \ib & InfiniBand \\
            \hbm & High-bandwidth memory \\
            \SA & Self-attention layer \\
            \MLP & Multi-layer Perceptron layer \\
            \LN & LayerNorm layer \\
            \LA & Logit-Attend layer \\
            \SM & Softmax layer \\
            DP & Data parallelism \\
            TP & Tensor parallelism \\
            2D TP & 2D tensor parallelism \\
            2D TP SUMMA & 2D tensor parallelism with SUMMA matrix-multiply \\
            PP & Pipeline parallelism \\
		% \hline
	\end{tabular}
 \caption{\emph{\revision{Abbreviations used in this paper.}}}
 \label{table:abbrev}
\end{table}

\noindent We now outline some additional details regarding the performance model. 

% \medskip \noindent \textbf{(S1) Counting FLOPs and bytes. }
% Most transformer operations can be distilled to the matrix multiply $\mathbf{C} = \mathbf{A}\mathbf{B}$ where $\mathbf{C} \in \mathbb{R}^{m \times n}, \mathbf{A} \in \mathbb{R}^{m \times k}, \mathbf{B} \in \mathbb{R}^{k \times n}$. The total FLOPs in this operation is $\lambda_f = (2k - 1)mn$. The memory accessed from \hbm ~is $\lambda_m = 2(mk + kn + mn)$ bytes (corresponding to each tensor), assuming FP16 (we also assume mixed precision training in our model). Assuming peak hardware FLOPs $\lambda_{fh}$ and peak hardware memory bandwidth from \hbm ~as $\lambda_{mh}$, roofline performance dictates peak performance time for the operation as $\max (\lambda_{f}/\lambda_{fh}, \lambda_{m}/\lambda_{mh})$. We can add communication time based on $V$ bytes of communication volume $t_\text{comm}$ (see \S\ref{sec:methods}) to this time to get the final estimate of the operation time. Depending on the operation, $t_\text{comm}$ may be overlapped with compute or exposed. Similar expressions can be derived for \LN, \SM, GELU, and Dropout. When operations are fused, as in \LA ~where the attention matrix ($\mathbf{A}$), \SM ~and attend operation to compute $\mathbf{S}$ are fused to a single operation, the amount of bytes counted only depend on the inputs to the fused operation and no intermediates (which increases the arithmetic intensity).

\medskip \noindent \textbf{(S1) Backward pass. } 
Assuming the matrix multiply, $\mathbf{C} = \mathbf{A}\mathbf{B}$ where $\mathbf{C} \in \mathbb{R}^{m \times n}, \mathbf{A} \in \mathbb{R}^{m \times k}, \mathbf{B} \in \mathbb{R}^{k \times n}$,
the backward pass computes the gradient tensors of the loss function $\mathcal{L}$:
\begin{align*}
    \frac{\partial \mathcal{L}}{\partial \mathbf{A}} &= \frac{\partial \mathcal{L}}{\partial \mathbf{C}} \mathbf{B}^T, 
    \frac{\partial \mathcal{L}}{\partial \mathbf{B}} = \mathbf{A}^T \frac{\partial \mathcal{L}}{\partial \mathbf{C}} 
\end{align*}
Note that $\mathbf{A}, \mathbf{B}$ are intermediate maps stored for the backward pass in \hbm. $\mathbf{B}$ may be weights (in MLPs) or activation maps (in \LA). Similar FLOPs, bytes and communication volumes can be derived for the backward pass (typically incurring twice the cost of the forward pass).

\medskip \noindent \textbf{(S1) 2D TP SUMMA. }
Scalable Universal Matrix Multiplication (SUMMA) \cite{van1997summa} is an efficient distributed matrix multiply algorithm that uses a 2D grid of processors (GPUs) $n_1 \times n_2$.
% In this algorithm, matrix-matrix multiplication is expressed  as a sum of outer products of sub-blocks of the original matrices.
\begin{algorithm}  
	\caption{$\mathbf{C} = \mathbf{A}\mathbf{B}$ using SUMMA}
	\label{algoC=AB}
	\begin{algorithmic}[1]
		\State \textbf{Input}: $\mathbf{A}_{ij}$, $\mathbf{B}_{ij}$
		\State \textbf{Output}: $\mathbf{C}_{ij}$
		\State $\mathbf{C}=0$
		\For{$\kappa = 0 \to n_b-1$}  
		\State \textbf{for} $i = 0,...,n_1-1$ Broadcast $\mathbf{A}^{\kappa}_i$ to $i^{th}$ process row
		\State \textbf{for} $j = 0,...,n_2-1$ Broadcast $\mathbf{B}^{\kappa}_j$ to $j^{th}$ process col
		\State $\mathbf{C}_{ij}=\mathbf{C}_{ij}+\mathbf{A}^{\kappa}_i\mathbf{B}^{\kappa}_j$
		\EndFor  
		\State \Return{$\mathbf{C}_{ij}$}  
	\end{algorithmic}  
\end{algorithm}
While many implementations of SUMMA \cite{Li:2023} assume square matrices and process grids, SUMMA does not impose these restrictions and we outline the general algorithm here. Each matrix is evenly partitioned into $n_1\times n_2$ sub-blocks and mapped onto their respective GPUs.
In the rectangular version of SUMMA \cite{lewis2022large}, we assume that $k$ is further divided into $n_b$ ``panels'', each of size $k_b$. Denoting the panels as $\mathbf{A}^\kappa = \mathbf{A}_{0:m-1,k':k''}, \mathbf{B}^\kappa = \mathbf{B}_{k':k'',0:n-1}$ with $k' = \kappa k_b$ and $k'' = (\kappa + 1)k_b - 1$, SUMMA recasts the matrix multiply as:
\begin{equation*}
    \mathbf{C} = \sum_{\kappa}^{n_b-1}\mathbf{A}^\kappa \mathbf{B}^\kappa.
\end{equation*}
% $A_{ij}$ ($i,j=0,1,...n-1$) denotes the sub-block of $A$ in $i$'s row and $j$'s column. 
Assuming the matrices are distributed as $\mathbf{A}_{ij}, \mathbf{B}_{ij}, \mathbf{C}_{ij}$, with $i = 0,1,...,n_1-1, j = 0,1,...,n_2-1$, and $\mathbf{A}^\kappa_i$ denotes panel at the $i^{th}$ row and $\mathbf{B}^\kappa_j$ denotes panel at the $j^{th}$ column, we show the SUMMA pseudo-code in Algorithm \ref{algoC=AB}.
%\ref{algoC=ABT} and \ref{algoC=ATB}.
The pseudo-code of $\mathbf{C} = \mathbf{A}^T\mathbf{B}$ and $\mathbf{C} = \mathbf{A}\mathbf{B}^T$ are similar but have a Broadcast and Reduce collective, instead of two Broadcasts (see \cite{van1997summa} for more details). 
% \begin{algorithm}
% 	\caption{$C=AB^T$}
% 	\label{algoC=ABT}
% 	\begin{algorithmic}[1]
% 		\State \textbf{Input}: $A_{ij}$, $B_{ij}$
% 		\State \textbf{Output}: $C_{ij}$
% 		\State $C_{ij}=0$
% 		\For{$l=0 \to n-1$}
% 		\State broadcast $B_{lj}$ within any column
% 		\State $C^{temp}_{ij}=A_{ij}(B_{lj})^T$
% 		\State reduce $C^{temp}_{ij}$ within any row to $C_{il}$
% 		\EndFor
% 		\State \Return{$C_{ij}$}
% 	\end{algorithmic}
% \end{algorithm}

% \begin{algorithm}  
% 	\caption{$C=A^TB$}
% 	\label{algoC=ATB}
% \begin{algorithmic}[1]
% 		\State \textbf{Input}: $A_{ij}$, $B_{ij}$
% 		\State \textbf{Output}: $C_{ij}$
% 		\State $C_{ij}=0$
% 		\For{$l = 0 \to n-1$}  
% 		\State broadcast $A_{il}$ within any row
% 		\State $C^{temp}_{ij}=(A_{il})^TB_{ij}$
% 		\State reduce $C^{temp}_{ij}$ within any column to $C_{lj}$
% 		\EndFor  
% 		\State \Return{$C_{ij}$}  
% 	\end{algorithmic}  
% \end{algorithm}
\begin{table}[htbp]
  \centering
  \rowcolors{2}{gray!25}{white} % Alternating row colors
  \begin{tabular}{c|c|c|c}
    % \hline
    Operation & Partitioned Tensor Shapes & Type &  Vol \\
    \hline
    \multicolumn{4}{|c|}{\textbf{2D TP with SUMMA over $n_1 \times n_2$ grid of GPUs}} \\
    \hline
    \multicolumn{4}{c}{\textit{SA}} \\
    \hline
    $\tilde{\mathbf{X}} = \text{LN}(\mathbf{X})$   & $\tilde{\mathbf{X}}: (b,\frac{l}{n_2},\frac{e}{n_1})$, $\mathbf{X}: (b,\frac{l}{n_2},\frac{e}{n_1})$,  & $\mathcal{AR}$ & $b\frac{l}{n_2}e$ \\
    $\mathbf{Q} = \tilde{\mathbf{X}} \mathbf{W_Q}$   & $\mathbf{Q}: (b,\frac{h}{n_1},\frac{l}{n_2},e_h)$, $\mathbf{W_Q}: (\frac{e}{n_2},\frac{e}{n_1})$,  & $\mathcal{B}$ & $V_1$ \\
    $\mathbf{A}=\mathbf{Q} \mathbf{K}^T$   & $\mathbf{A}: (b,\frac{h}{n_1},\frac{l}{n_2},l)$, $\mathbf{K}: (b,\frac{h}{n_1},l,e_h)$  & $\mathcal{AG}$ & $bl\frac{e}{n_1}$ \\
    $\mathbf{S} = \mathbf{A}\mathbf{V}$   & $\mathbf{S}: (b,\frac{h}{n_1},\frac{l}{n_2},e_h)$, $\mathbf{V}: (b,\frac{h}{n_1},l,e_h)$  & $\mathcal{AG}$ & $bl\frac{e}{n_1}$ \\
    $\mathbf{Y} =  \mathbf{S}\mathbf{W_p}$   & $\mathbf{Y}: (b,\frac{l}{n_1n_2},e)$, $\mathbf{W_p}: (\frac{e}{n_1},e)$  &  $\mathcal{RS}$ & $b\frac{l}{n_2}e$ \\
    \hline
    \multicolumn{4}{c}{\textit{MLP}} \\
    \hline
    $\tilde{\mathbf{Y}} = \text{LN}(\mathbf{Y})$   & $\tilde{\mathbf{Y}}: (b,\frac{l}{n_2},\frac{e}{n_1})$, $\mathbf{Y}: (b,\frac{l}{n_2},\frac{e}{n_1})$,  &  $\mathcal{AR}$ & $b\frac{l}{n_2}e$ \\
    $\mathbf{Z} = \tilde{\mathbf{Y}}\mathbf{W_1}$   & $\mathbf{Z}: (b,\frac{l}{n_2},\frac{f}{n_1})$, $\mathbf{W_1}: (\frac{e}{n_2},\frac{f}{n_1})$  & $\mathcal{B}$ & $V_2$ \\
    $\mathbf{X} = \mathbf{Z} \mathbf{W_2}$   & $\mathbf{X}: (b,\frac{l}{n_2},\frac{e}{n_1})$, $\mathbf{W_2}: (\frac{f}{n_2},\frac{e}{n_1})$  &  $\mathcal{B}$ & $V_3$\\
    % \hline
    % \multicolumn{4}{|c|}{\textbf{2D: TP}} \\
    % \hline
    % % \hline
  \end{tabular}
  \caption{\emph{2D TP SUMMA: Tensor shapes, communication collective and volume (vol: total bytes transferred per GPU) for different operations. $\mathcal{AG}$ is AllGather, $\mathcal{RS}$ is ReduceScatter, $\mathcal{B}$ is Broadcast, and $\mathcal{AR}$ is AllReduce. $\mathbf{K}, \mathbf{V}$ follow $\mathbf{Q}$. Communication volume scales with one GPU dimension and there are no shared weights. For the SUMMA multiplies, $V_1 = ble/n_2 + e^2/n_1, V_2 = V_3 = ble/n_2 + ef/n_1$.}}
  \label{tab:2D-summa}
\end{table}

\begin{figure}
    \centering
    \includegraphics[width=0.8\linewidth]{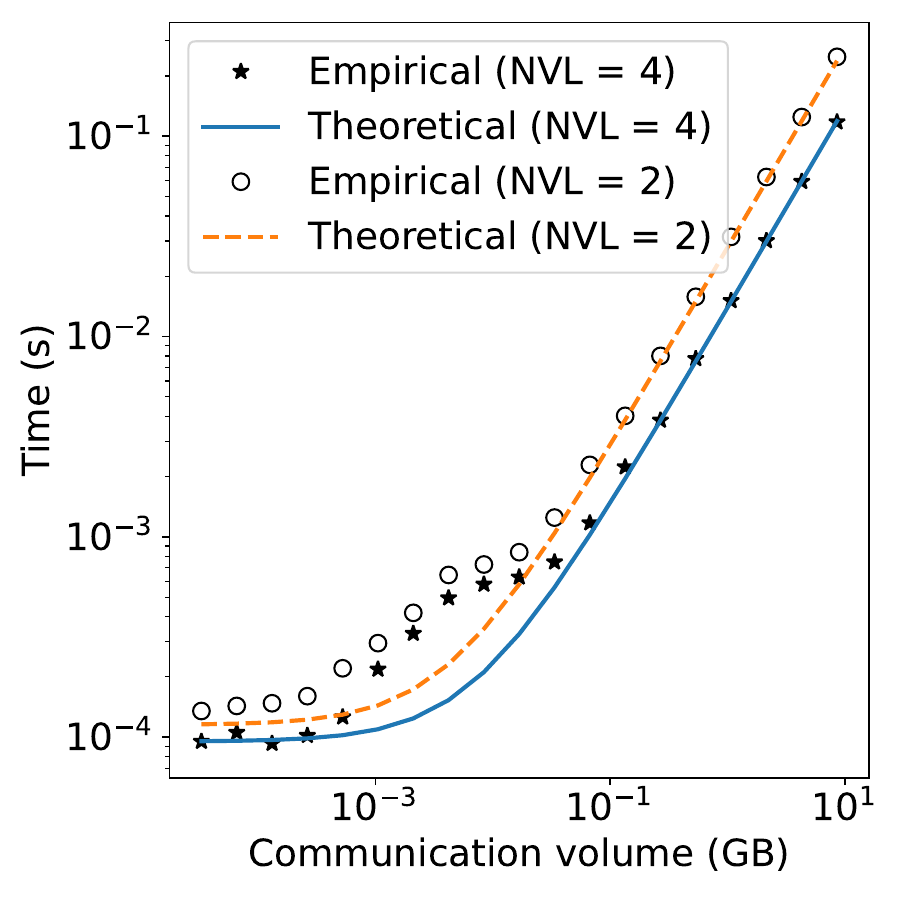}
    \caption{\emph{Time for $\mathcal{AG}$ as a function of communication volume on 32 \amp ~GPUs. We compare empirical numbers on Perlmutter through NCCL tests \cite{nccltests} to our theoretical formulae for different fast domain (NVLink) sizes. For NVL 2, 2 GPUs per node are used and for NVL 4, 4 GPUs per node are used. We see that more GPUs per node effectively increases the SlingShot bandwidth, leading to smaller times.}}
    \label{fig:comm-validation}
\end{figure}

\medskip \noindent \textbf{(S2) Communication in TP. }
We show the full SUMMA communicaton and tensor shapes in Tab. \ref{tab:2D-summa}. Note that $V_1, V_2, V_3$ all communicate the weights as well as the activation maps, increasing the absolute volume but showing better scaling with $n_1, n_2$.
For the forward pass, in 1D TP, the communications outlined in Tab. \ref{tab:1D} are assumed to be non-overlapped since partial sums are assumed to be computed before the communication can take place and the subsequent layers need to wait for the synced activation map. 2D TP is similar with the $l$ partitions being embarrassingly parallel. In SUMMA, we assume some overlap of the TP communications. There is a prologue time $t_{\text{prologue}}$ to perform two broadcasts before the first $\kappa$ iteration and the rest of the broadcasts (for subsequent $\kappa$ iterations) can be scheduled asynchronously (and hence overlapped) to the computation of $\mathbf{C}_{ij}$. Hence, the communication time $t_\text{comm} = t_{\text{prologue}} + n_b t_{\text{exposed}}$, with $t_{\text{exposed}}$ as the exposed time between two broadcasts and the compute. Hence, the $n_b$ parameter can influence $t_\text{comm}$ as larger values will reduce $t_{\text{prologue}}$ (smaller panels communicated) but at the same time reduce the matrix multiply efficiency (smaller matrices, may become memory bound with small inner dimensions), exposing communication more. Hence, the 2D TP configuration also searches over $n_b$.
In the backward pass, 1D TP incurs similar $\mathcal{AG}$ and $\mathcal{RS}$ in a conjugate fashion to the forward pass communications due to the transposed matrix multiplies. In 2D TP, the weight gradients incur an additional reduction across $n_2$ due to the partitioned $l$ dimension. We assume this TP communication to be overlapped and scheduled with the DP communications---\revision{the weight gradient $\mathcal{RS}$ and the weights $\mathcal{AG}$ happen across the combined $n_d \times n_2$ GPU group}. For SUMMA, the transposed matrix multiples incur a Broadcast and Reduce instead of two Broadcasts.

\begin{table}
  \centering
  \rowcolors{2}{gray!25}{white} % Alternating row colors
	\begin{tabular}{l |c |c | c } 
		% \hline
		Description & \amp & \hop  & \blackwell \\ 
		\hline
		Tensor core FP16 (TFLOPs/s) & 312 & 990 & 2500  \\ 
		% \hline
		Vector FP16 (TFLOPs/s)& 78& 134 & 339    \\
		% \hline
		% Vector Flops FP16 & Half precision matrix-vector multiplication  [TFlops/s]&  78& 133.8 & 339  \\
		% \hline
		Flops Latency (s) & 2e-5 & 2e-5& 2e-5 \\
		% \hline
		HBM Bandwidth (GB/s) & 1555 & 4800 & 8000  \\
		% \hline
		HBM Capacity (GB) & 80 & 141 & 192  \\ 
		% \hline
		\nvs ~1-directional Bandwidth (GB/s) & 300 & 450 & 900  \\
		% \hline
		% NVLink Efficiency & Bandwidth efficiency & 0.7& 0.7& 0.7  \\
		% \hline
		\nvs ~Latency (s) & 2.5e-6 & 2.5e-6& 2.5e-6\\
		% \hline
		\ib ~Bandwidth (GB/s) & 25 & 50 &  100\\
		% \hline
		% IB Efficiency& Efficiency of Infiniband bandwidth  [GB/s] & 0.7& 0.7&  0.7 \\
		% \hline
		\ib ~Latency (s) & 5e-6 & 5e-6 & 5e-6\\
		% \hline
	\end{tabular}
 \caption{\emph{GPU and network parameters for various GPU generations.}}
 \label{table:GPU_generation_params}
\end{table}
% 2D TP rationales

\begin{figure*}[t]
    \centering
    \begin{subfigure}[b]{0.49\textwidth}
        \centering
        \includegraphics[width=\textwidth]{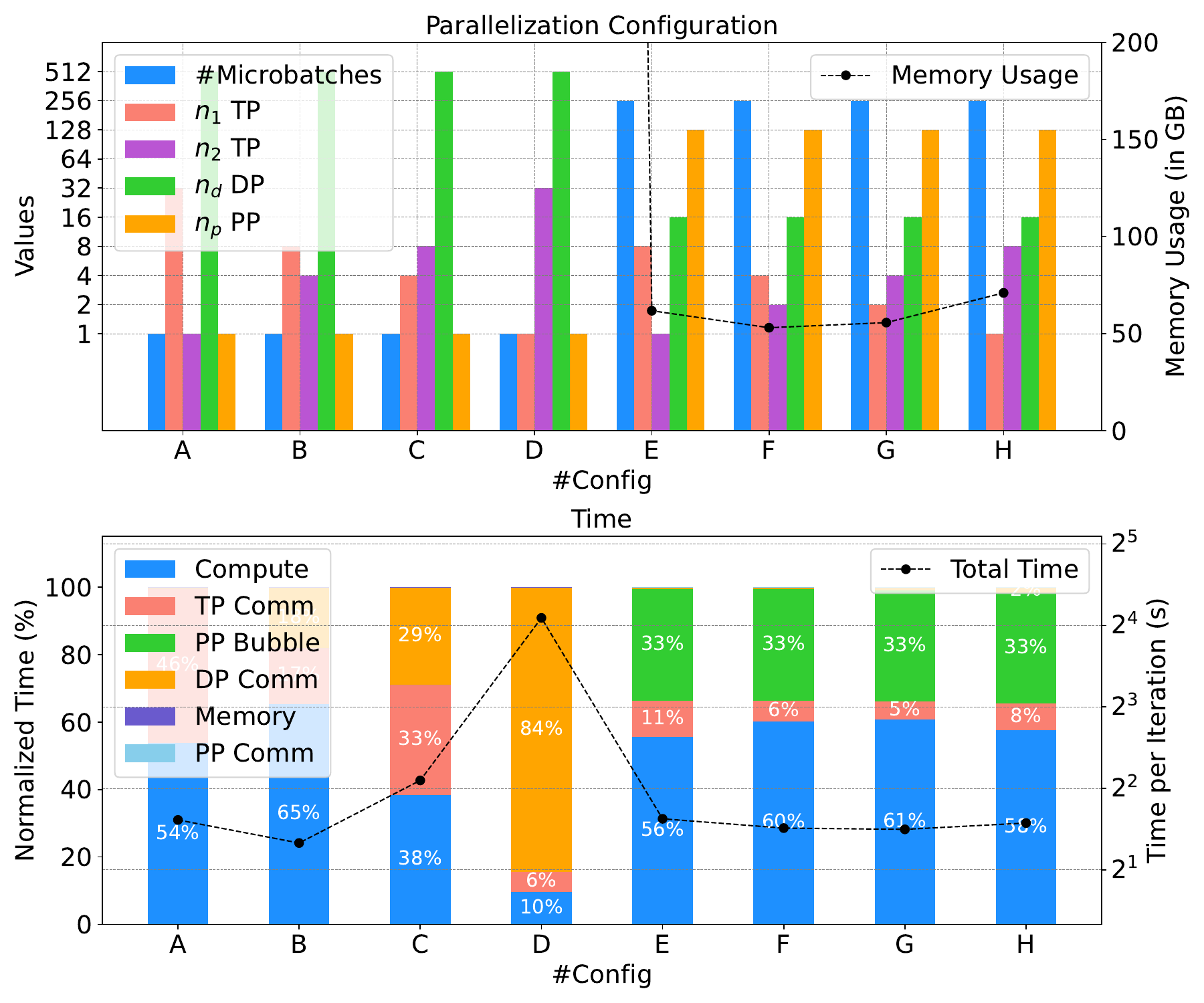}
        \caption{\textbf{\gpt ~with 2D TP}}
        \label{fig:gpt3-2d-tp}
    \end{subfigure}%
    % \hfill
    \begin{subfigure}[b]{0.49\textwidth}
        \centering
        \includegraphics[width=\textwidth]{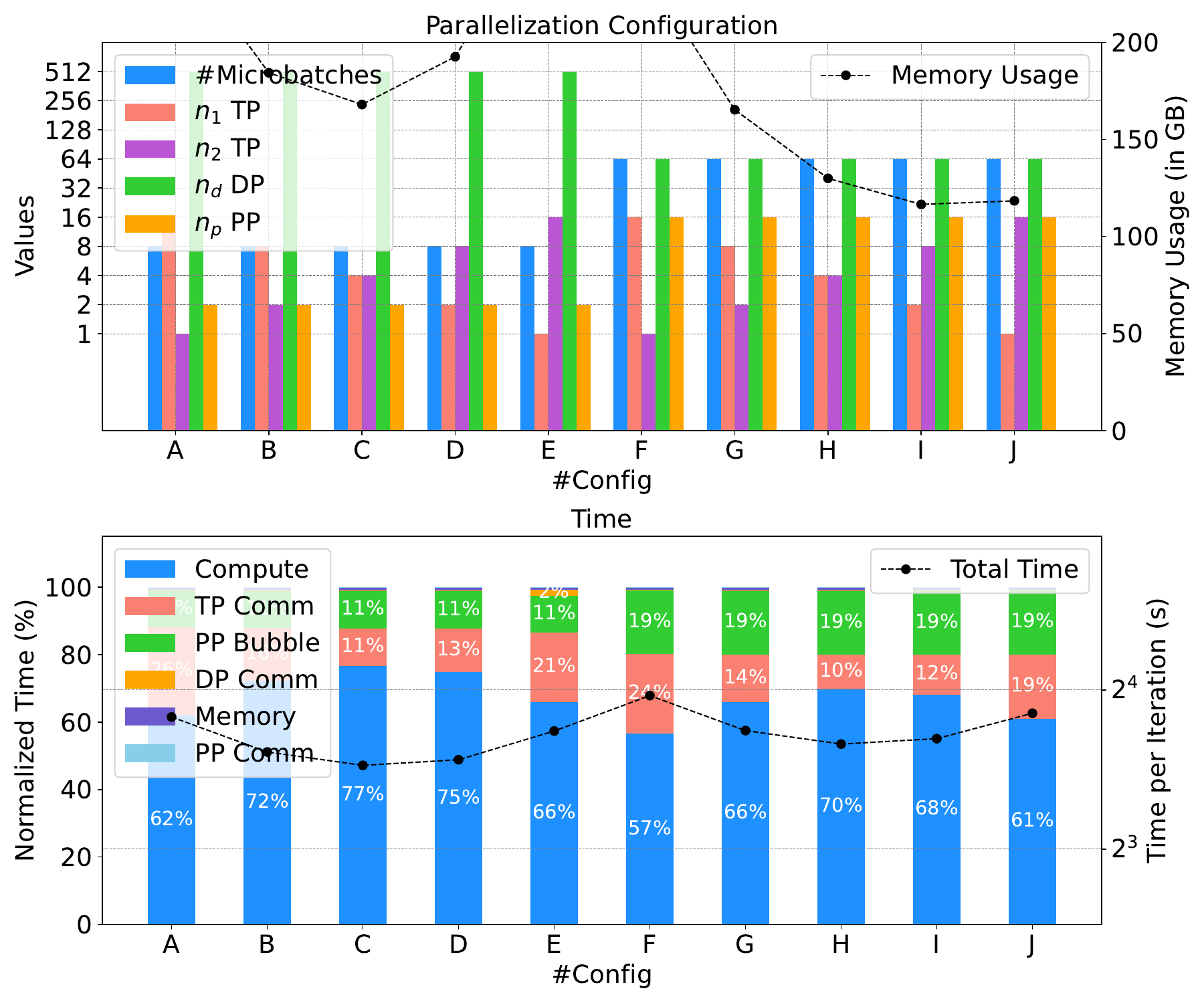}
        \caption{\textbf{\vit ~with 2D TP}}
        \label{fig:vit-2d-tp}
    \end{subfigure}
    \caption{\emph{(a) For \gpt ~with 2D TP, we fix TP $n_t=32, n_p=1$ and vary $n_1, n_2$ to get the the first five configurations and then switch to $n_p = 128$ and repeat the same with $n_t=8$ for the last four configurations. We use a \blackwell ~GPU with \nvs ~domain size $n_{\text{\nvs}}=64$, global batch size $4096$ on $16384$ GPUs and also show time for each configuration broken down by the components. We see similar behavior as 2D TP SUMMA but the memory consumed is very high and hence the large PP configurations are chosen. (b) For \vit ~with 2D TP, we first fix $n_t=16$ and vary $n_1,n_2$, then switch to $n_p=16$ and repeat the same. The memory used is sensitive to the parallelization regime.}}
    \label{fig:2d-tp-rationale}
\end{figure*}

\medskip \noindent \textbf{(S2) Validating communication time of collectives. }
We validate our time to communication formulae for the different communication collectives on the Perlmutter~\cite{perlmutter} supercomputer at NERSC. Perlmutter has 4 \amp ~GPUs per node connected via 4 third generation NVLinks between each pair of GPUs (with all GPUs interconnected) and 4 NICs per node for the SlingShot communications (similar to \ib). While there is no NVSwitch on Perlmutter, we can derive equivalent expressions for NVLink based on number of NVLinks per GPU used in the collective. For example, if only 2 GPUs per node were used, only 4 NVLinks are used for the fast bandwidth. If all 4 GPUs are used in a node, then 12 NVLinks are used. In Fig. \ref{fig:comm-validation}, we show that our theoretical time formulae agree well with the empirical results. We also see the effect of the increased SlingShot bandwidth if more GPUs are used per node (in the fast domain). We observe some non-linear latency effects at small volumes and do not model these in our performance model. We also verify the above agreement over a range of GPUs (and nodes), and also for other collectives.

\medskip \noindent \textbf{(S2) Hardware configurations. }
We summarize the hardware parameters used in our model. The specifications for the  \amp ~GPU can be found in~\cite{A100}, and the \hop ~and \blackwell ~GPUS can be found respectively in ~\cite{H200} and~\cite{B200}. Each GPU generation is coupled to its respective NVLink generation. Similarly, the \ib ~interconnects improve with each generation, from ConnectX-6~\cite{IB6}, to ConnectX-7~\cite{IB7}, and the most recent ConnectX-8~\cite{IB8}. We assume network latency values from the previous validation experiments and also assume they do not greatly improve with newer generations. 
% In addition, here we also provide projections for a future GPU based on extrapolations from current GPUs. T
% 
% The values for bandwidth and speed compute are the maximal values the hardware can reach. 
In our experiments on Perlmutter, we observe typical bandwidth efficiencies of 70\% for the networks and include this as an efficiency parameter as well. For FLOPs, we assume a simple model based on \cite{matmul}: $t_\text{flops} = t_\text{sf} + \lambda_{f}/\lambda_{fh}$, where the FLOPs latency $t_\text{sf}$ can be inferred from \cite{matmul} and models the inefficiency in matrix multiplies of small matrices to the first order.

\medskip \noindent \textbf{Software environment setup for validation. }
We validate our performance model in \ref{sec:results} with the open source \textsc{Megatron-LM} codebase \cite{megatroncolde}. We use NVIDIA PyTorch NGC container \textsc{v24.05} and build \textsc{Megatron-LM} in the container. 
% do we mention the patch? i've been using a version off of trunk
% not GPT3-1T
The GPT3-175B and custom~\vit ~models are built on top of \textsc{TransformerEngine} and \textsc{FlashAttention-2}.
For optimized collective operations, NCCL v2.19.4 and its associated plugin for Perlmutter is used. Iteration times are captured using the built-in logging utility in \textsc{Megatron-LM}.
 We use the CodeParrot \cite{codeparrot} dataset for empirical validation experiments. This dataset is pre-processed by existing tools provided in Megatron-LM.
% NCCL backend is used for PyTorch Distributed initialization and the validation runs are launched with \texttt{srun}.    
The following optimizations are enabled during training: distributed optimizer, overlapping gradient reduction, overlapping parameter all-gather in the distributed optimizer, half precision (FP16) along with Flash Attention. 
To test different strategies that require mapping different ranks on different devices on and across nodes, we tweak the model parallel initialization to accept different mapping orders in the form of strings, for example, \texttt{tp-cp-dp-pp}, \texttt{cp-tp-pp-dp}, etc. 

% one liner about rank mapping as well

\subsection{Additional Results}
\label{sec:appendix_additional_results}
%We describe some additional results here. 
\begin{figure*}[t]
    \centering
    \begin{subfigure}[b]{0.48\textwidth}
        \centering
        \includegraphics[width=\textwidth]{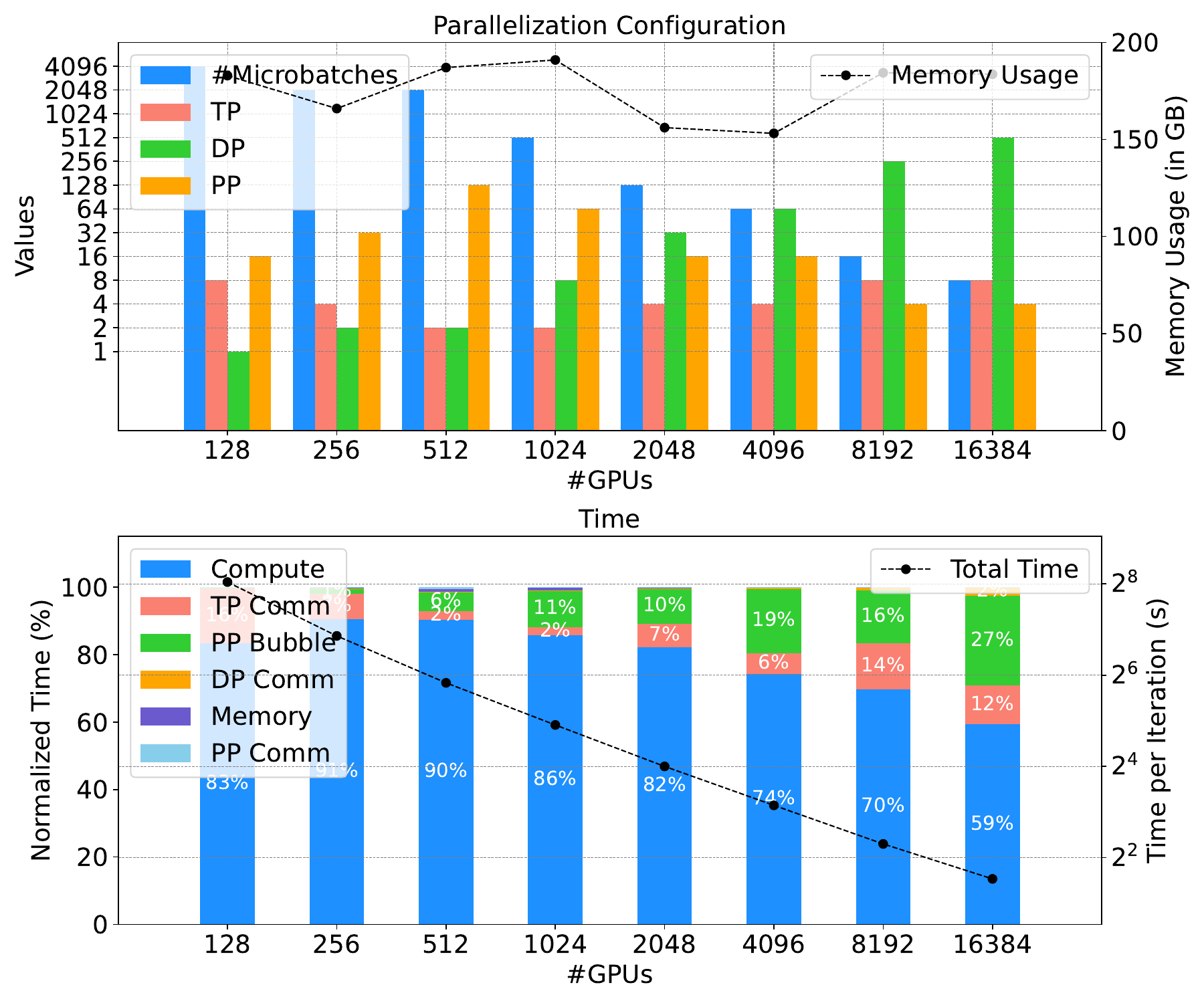}
        \caption{\textbf{\gpt ~1D TP on} $n_\text{\nvs} = 64$}
        \label{fig:gpt3-1d-tp-large-nvs}
    \end{subfigure}%
    % \hfill
    \begin{subfigure}[b]{0.48\textwidth}
        \centering
        \includegraphics[width=\textwidth]{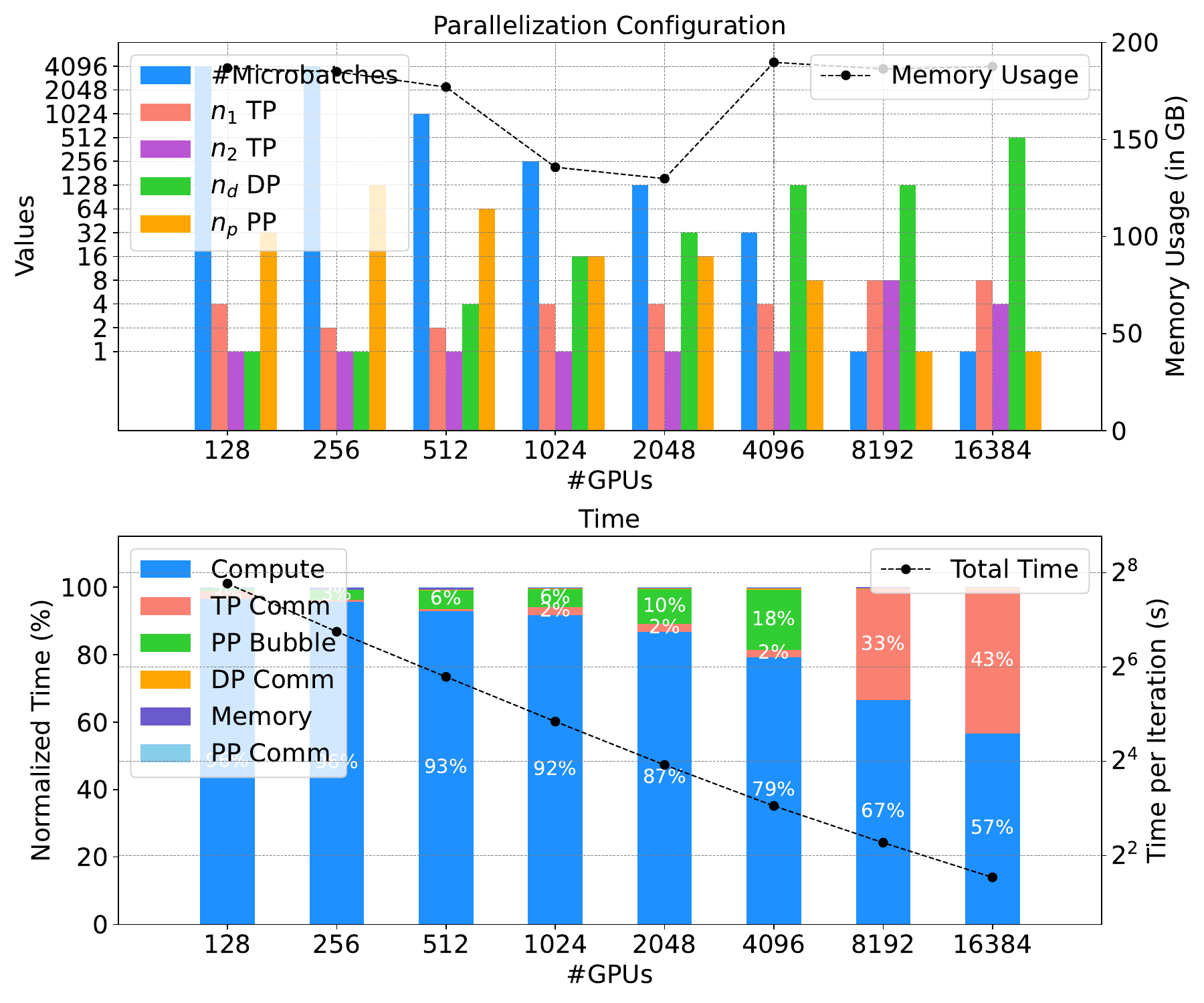}
        \caption{\textbf{\gpt ~2D TP SUMMA on} $n_\text{\nvs} = 64$}
        \label{fig:gpt-2d-tp-summa}
    \end{subfigure}
    \caption{\emph{On \blackwell ~GPU with global batch size $4096$: (top) Optimal parallelization strategy and \hbm ~memory consumed vs number of GPUs (bottom) Time vs number of GPUs broken down by different components. (a) \gpt ~with 1D TP on larger \nvs ~domain shows reduced PP at scale (b) \gpt ~with 2D TP SUMMA shows mostly 1D TP except at scale due to the large \nvs ~domain.}}
    \label{fig:gpt-all}
\end{figure*}

\begin{figure*}[t]
    \centering
    \begin{subfigure}[b]{0.5\textwidth}
        \centering
        \includegraphics[width=\textwidth]{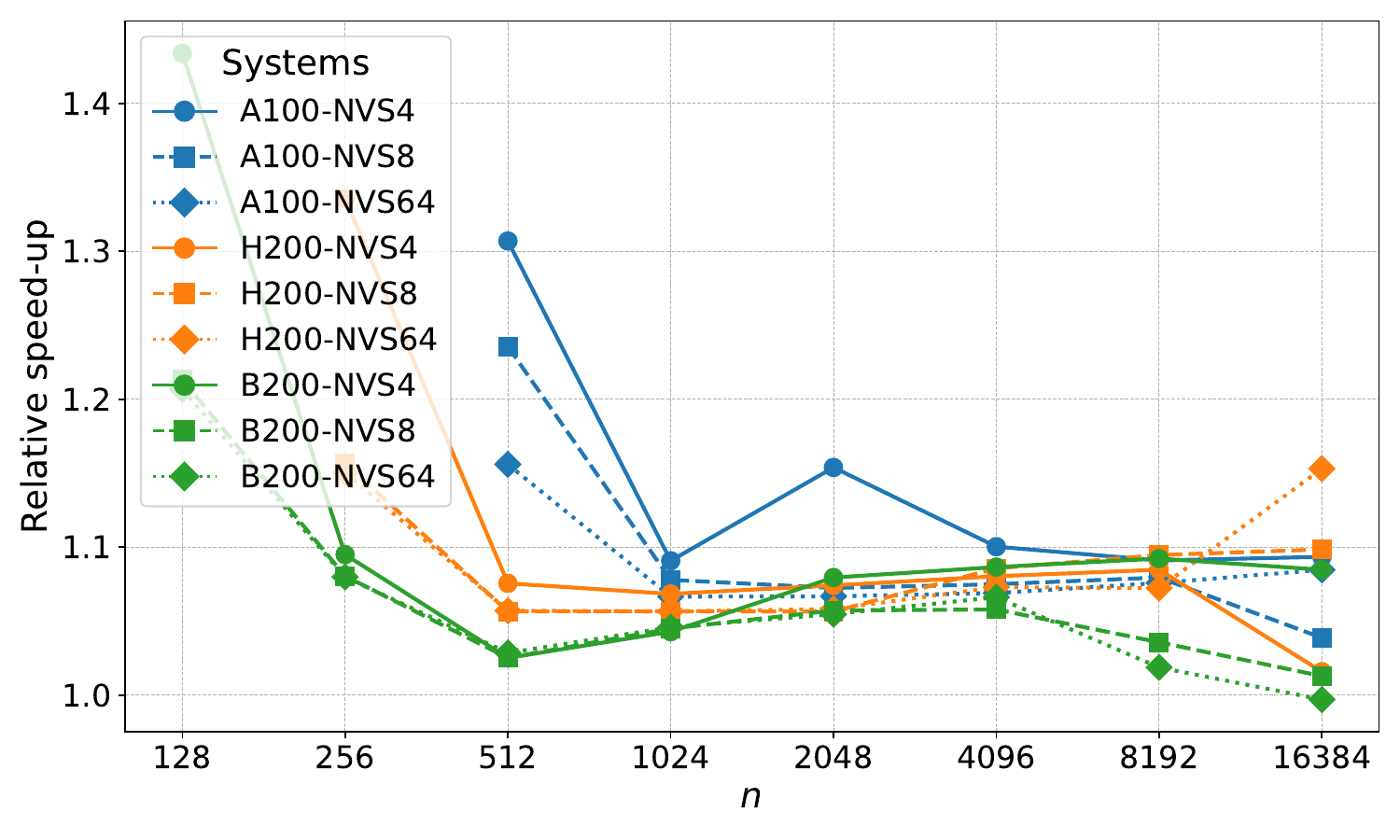}
        \caption{\textbf{Speedup for 2D TP SUMMA w.r.t 1D TP}}
        \label{fig:2d-1d}
    \end{subfigure}%
    % \hfill
    \begin{subfigure}[b]{0.5\textwidth}
        \centering
        \includegraphics[width=\textwidth]{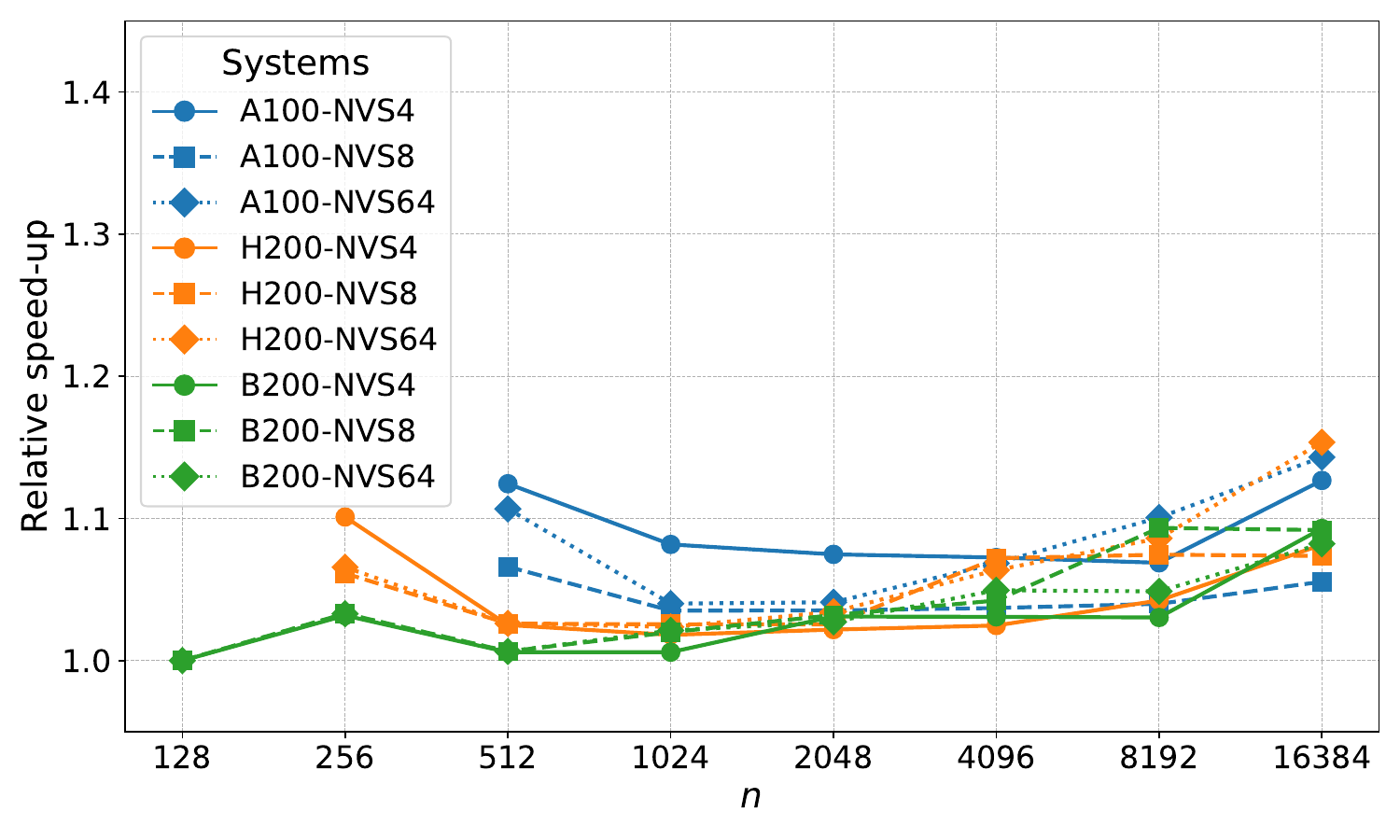}
        \caption{\textbf{Speedup for 2D TP w.r.t 1D TP}}
        \label{fig:2d-seqp-1d}
    \end{subfigure}
    \caption{\emph{We show the relative speedup in training time of the two 2D TP strategies with respect to 1D TP. The speedups are colored according to GPU generations and marked according to \nvs ~domain size. We observe that both versions provide good speedups of about 10\% with SUMMA particularly helpful at resource constrained regimes (small scale, capacity, \nvs ~domain size). 2D TP is more performant at the large scales. Higher GPU generations and \nvs ~sizes decrease speedup in general.}}
    \label{fig:speedup}
\end{figure*}

\medskip \noindent \textbf{2D TP rationale. } In Fig. \ref{fig:2d-tp-rationale}, we show that 2D TP works similarly to 2D TP SUMMA in \gpt ~with low PP/high DP and low DP/high PP configurations as possible candidates. However, unlike 2D TP SUMMA, the low PP/high DP configurations take up a lot of memory since 2D TP has lot of shared weights and activation maps compared to SUMMA. Also, the DP communication times are shown to be high, but in 2D TP, this includes the weight gradients reduction across $n_2$ that is scheduled with the DP reductions---we do not disentangle this TP communication in our estimates. For the \vit, we show the 2D TP possibilies in Fig. \ref{fig:vit-2d-tp}. We still observe the high and low PP configurations contending with each other (with high TP necessary for this model), with the low PP configurations being favored here. We also see that the memory usage is sensitive to the choice of $n_1$, $n_2$ with $n_p$. For small PP, the weights memory dominates at low $n_1$ and activation  memory dominates at low $n_2$. With high PP, the activation memory dominates and hence with more $n_2$, the memory comes down.

% ViT rationales
% GPT with NVS64 on b200
% ViT with 2D SUMMA

\begin{figure*}[t]
    \centering
    \begin{subfigure}[b]{0.5\textwidth}
        \centering
        \includegraphics[width=\textwidth]{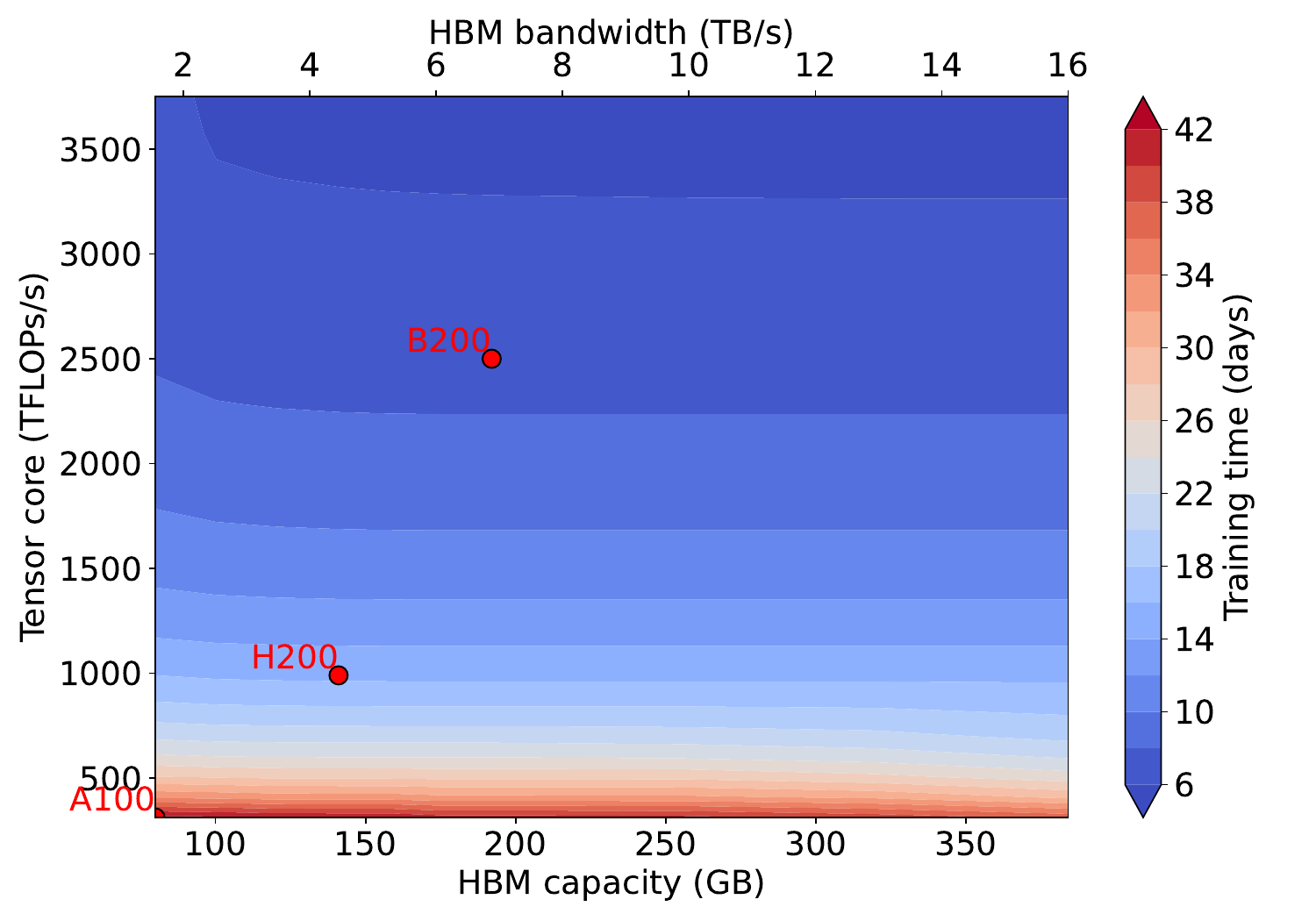}
        \caption{\textbf{\gpt ~with 1D TP}}
        \label{fig:gpt3-flops-hbm}
    \end{subfigure}%
    % \hfill
    \begin{subfigure}[b]{0.5\textwidth}
        \centering
        \includegraphics[width=\textwidth]{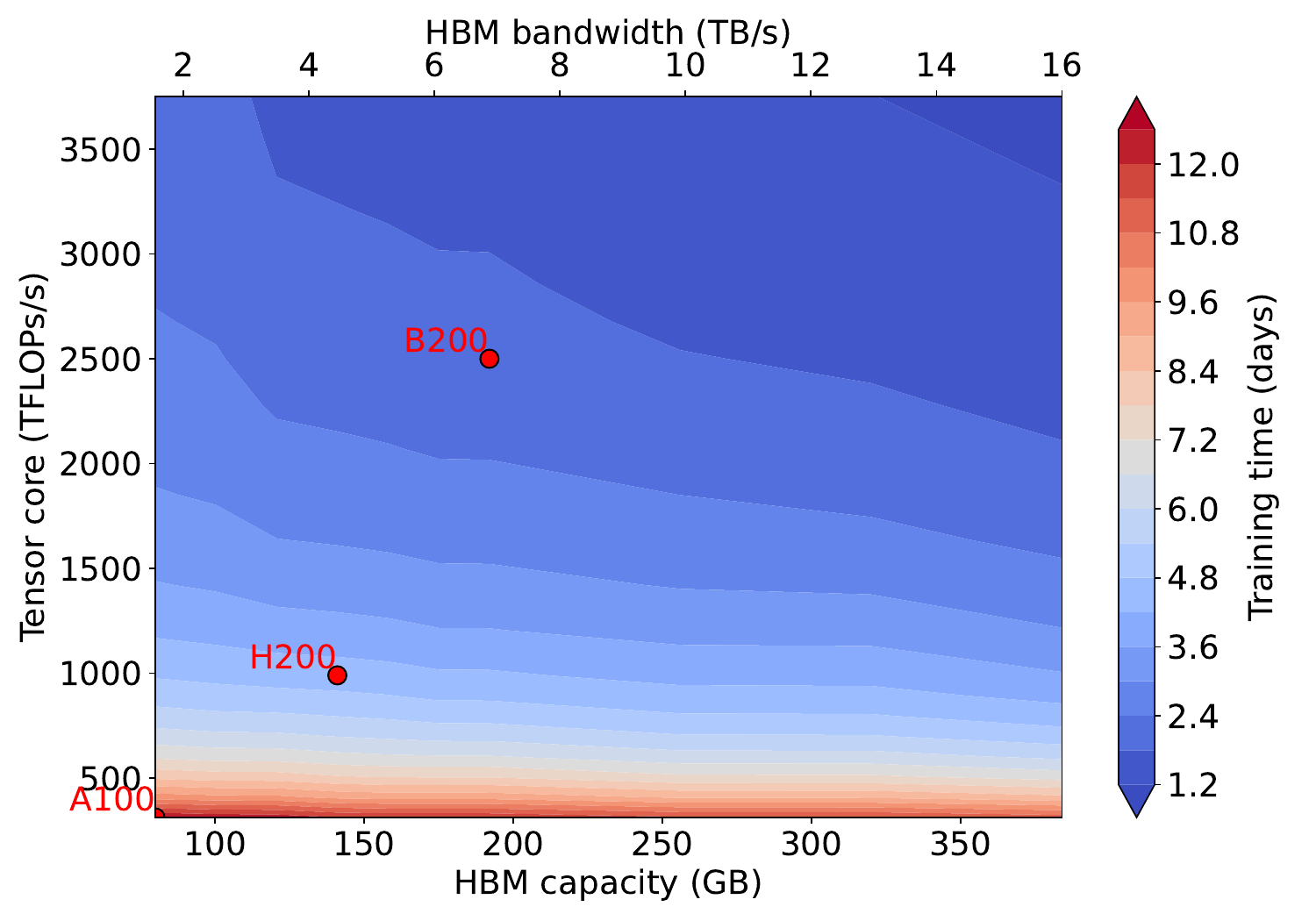}
        \caption{\textbf{\vit ~with 2D TP}}
        \label{fig:vit-flops-hbm}
    \end{subfigure}
    \caption{\emph{Training times for \gpt ~and \vit ~using 8192 GPUs as a function of FLOP rate (compute speed) and \hbm ~memory capacity and bandwidth. \nvs ~domain is 8 with fixed \nvs ~and \ib ~bandwidths at the \blackwell ~generation and the global batch size is $4096$. FLOP rates are the primary factor for speed-ups with bandwidth/capacity having different roles for the different models.}}
    \label{fig:flops-hbm}
\end{figure*}

\begin{figure*}
    \centering
    \begin{subfigure}{0.47\textwidth}
        \centering
        \includegraphics[width=\textwidth]{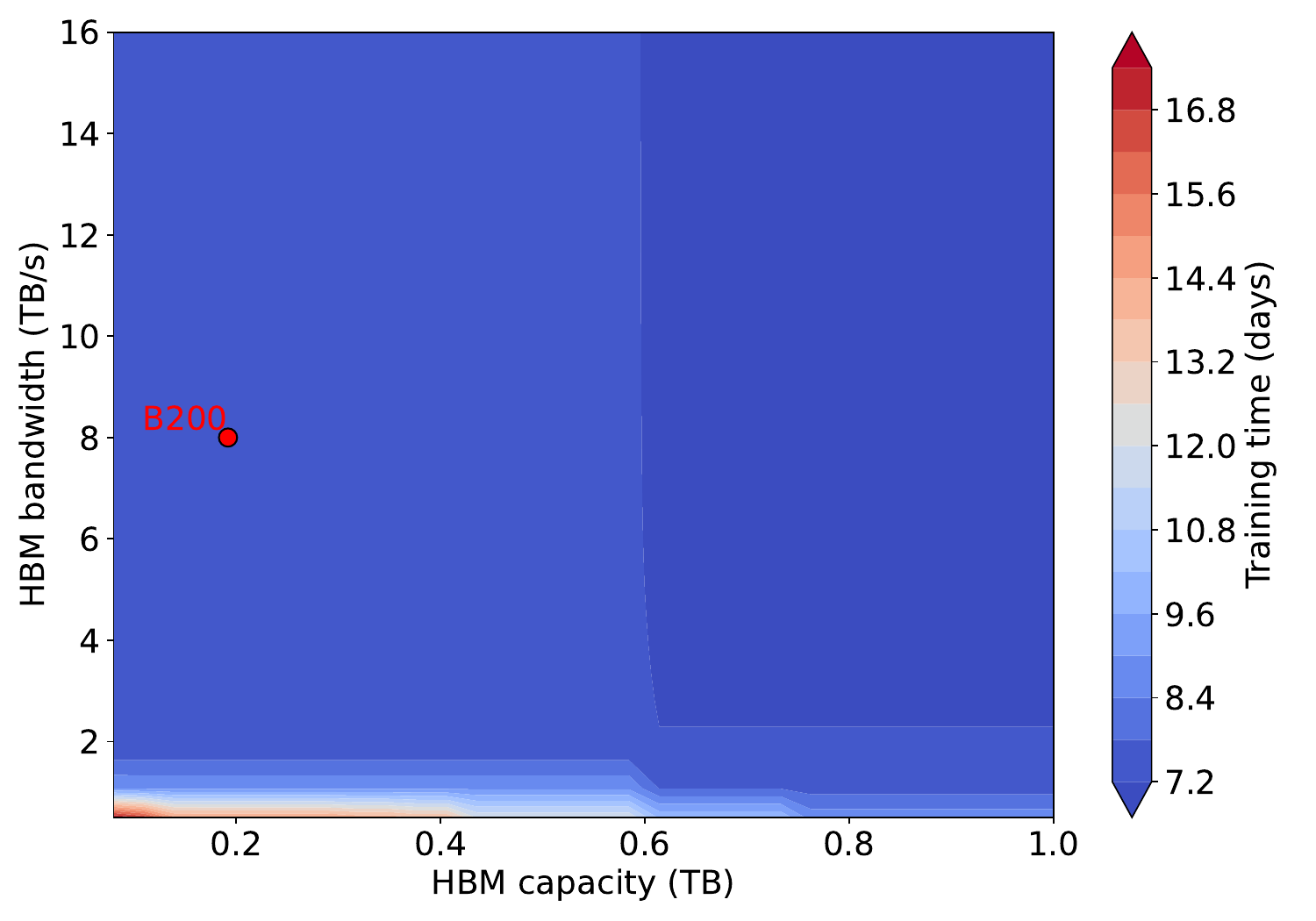}
        \caption{\textbf{\gpt ~with 1D TP}}
        \label{fig:gpt3-hbm}
    \end{subfigure}%
    % \hfill
    \begin{subfigure}{0.47\textwidth}
        \centering
        \includegraphics[width=\textwidth]{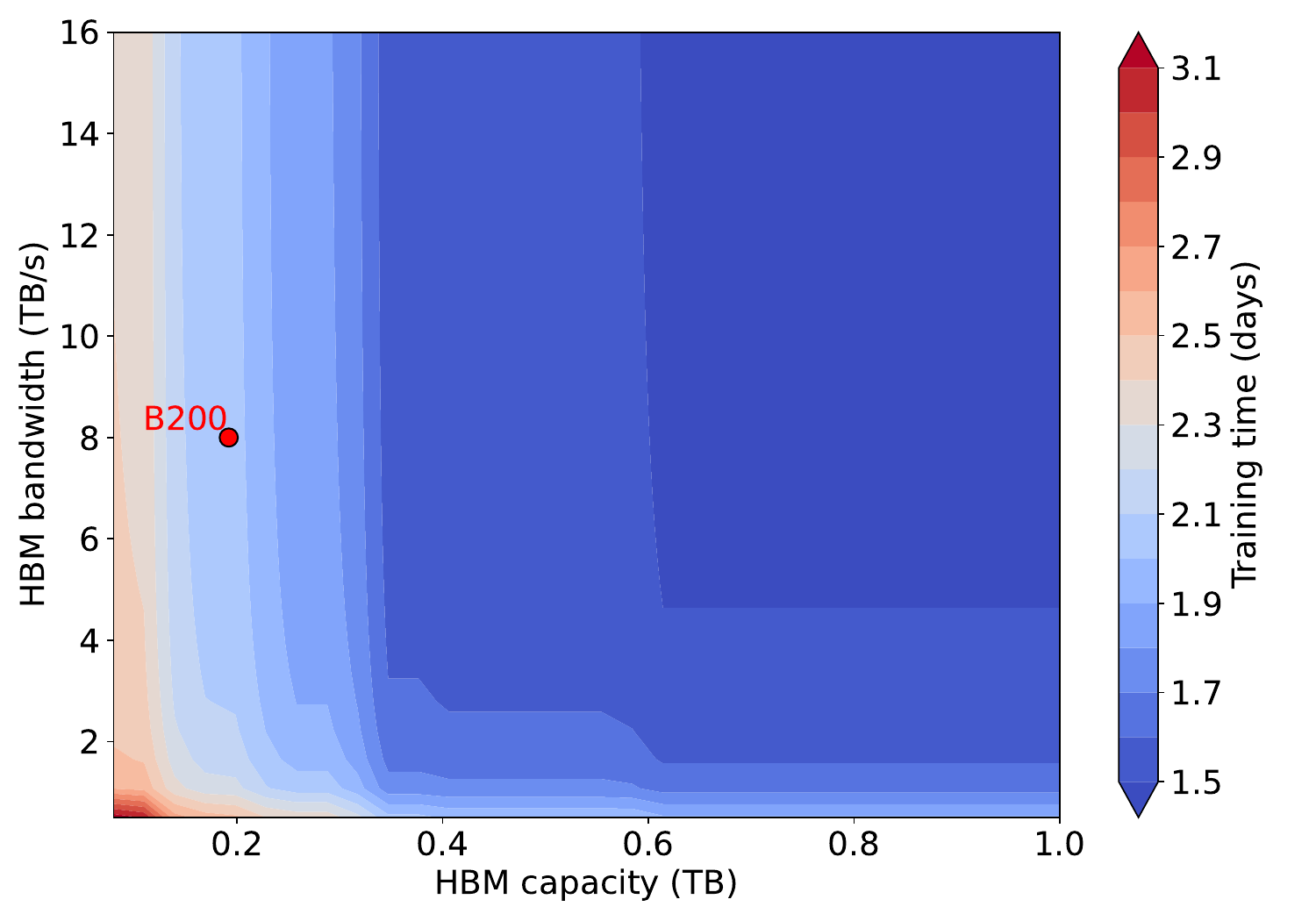}
        \caption{\textbf{\vit ~with 2D TP}}
        \label{fig:vit-hbm}
    \end{subfigure}
    \caption{\emph{Training time for \gpt ~and \vit ~as function of \hbm ~memory capacity and bandwidth. \nvs ~domain is 8 with fixed \nvs ~and \ib ~bandwidths at the \blackwell ~generation (along with the FLOP rates) and the global batch size is $4096$. Both models show lower bandwidth/higher capacity configurations can contend in performance by trading-off lesser parallelism inefficiencies for higher memory access times.}}
    \label{fig:hbm}
\end{figure*}

\medskip \noindent \textbf{Parallelization and model type. }
For \gpt, we show 1D TP on a large \nvs ~domain in Fig. \ref{fig:gpt3-1d-tp-large-nvs}. We see that small PP (and large DP) is favored at scale due to the large fast bandwidth domain. With 2D TP SUMMA, we see that, for \gpt, the model effectively chooses 1D TP at most scales. Due to the large \nvs ~domain, it chooses 2D partitioning at scale. For the \vit, the 2D TP SUMMA is very similar to 2D TP, with the latter showing better performance (training times) and the former showing better memory utilization.

\medskip  \noindent \textbf{Speedups with higher-dimensional TP. }
\revision{We show that both 2D TP versions for \gpt ~can reduce training time. In Fig. \ref{fig:speedup}, we show the speedup of the two 2D TP versions with respect to 1D TP for different GPU generations and \nvs ~domain sizes. While the speedups are generally clustered for most systems, we observe speedups of approximately  5--10\% across various GPU scales. 2D TP SUMMA is more helpful at very small scales, small capacities (\amp), and smaller \nvs. 
% In \amp, SUMMA also helps when \nvs ~size is small. 
2D TP shows similar speedups that are higher at the larger scale. }
% In general, both 2D versions show smaller speedups with higher GPU generations and larger \nvs ~sizes.}

\medskip  \noindent \textbf{GPU memory bandwidth, capacity, and FLOP rate effects. }
In Fig.~\ref{fig:flops-hbm}, we plot the training times for \gpt ~and \vit ~as function of GPU compute vs HBM memory and bandwidth. To understand the impact of the GPU parameters alone, we maintain the same network architecture while scaling up the memory capacity and bandwidth on one axis, and the tensor core and vector FLOP rates on the other. We show both memory capacity and bandwidth on the x-axis and only the tensor core FLOPs/s on the y-axis without loss of generality. Also, we display in the plots three generations of GPUs (\amp, \hop, \blackwell).
As observed in \S\ref{sec:results}, \gpt ~is less sensitive to HBM capacity and bandwidth at large scales, with tensor core FLOPs/s being the primary factor for performance boosts. For \vit, due to the large TP necessary for the long sequence model (see discussion in \S\ref{sec:results} around performance senstitivty to the model type), the capacity and bandwidth play a bigger role. 
%
% ViT with 2D parallelism shows higher dependence of memory capacity up to 300 GB, and changes slowly, with 500 GB and 900 GB being other major inflection points. A similar behavior is found for memory bandwidth with (300, 1300, 1500) TB/s as major points of throughput increase. Both models show dependence on memory, but for different reasons. The first model is very large and higher memory translates to lower model parallelism and communication. The second model is not as large but the input sequence puts high constraints on both compute and memory due to self-attention operations and activations memory footprint scaling quadratically and linearly with sequence length respectively.

%
In Fig.~\ref{fig:hbm}, we use the \blackwell ~generation compute and network speeds and vary separately the \hbm ~memory and bandwidth. We once again see the weak dependence on capacity and bandwidth for the \gpt ~model---only very small bandwidths increase memory-bound operation times. However, we observe that large capacities and small bandwidths, representative of different memory technologies like LPDDR \cite{jedec-lpddr5}, show good performance, comparable to the \blackwell. We see high-bandwidth/low-capacity configurations comparable to the low-bandwidth/high-capacity configurations, where lesser parallelism (and hence communication/bubbles) is traded off for larger memory access times.
The \vit ~once again shows more sensitivity to the capacity and bandwidth with multiple inflection points, with smaller capacities showing poorer performance. However, again the high capacity, low bandwidth regions show good performance compared to the \blackwell, highlighting alternate memory technologies may benefit training these models at certain scales. With more capacity, this can also help reduce the dependence of the \vit ~on \nvs ~as lesser TP is required to fit the model.
\end{document}